\documentclass[a4paper,fleqn]{cas-sc}

\usepackage[authoryear]{natbib}

\usepackage{tikz}
\usetikzlibrary{arrows.meta, positioning, shapes.geometric, backgrounds, fit, calc, decorations.pathmorphing}
\usepackage{xcolor}
\usepackage{hyperref}
\usepackage{cleveref}  
\usepackage[figuresright]{rotating}
\usepackage{float}
\usepackage{algorithm}
\usepackage{algpseudocode}
\usepackage[htt]{hyphenat}

\newcommand{\myshorttitle}{Dynamic Bidirectional Pattern Memory}
\newcommand{\myshortauthors}{A.H. Lazem et al.:}
\usepackage{fancyhdr}
\usepackage{lastpage}
\usepackage{pifont}  
\newcommand{\yes}{\textcolor{ForestGreen}{\ding{108}}}  
\newcommand{\no}{\textcolor{black!30}{\ding{109}}}       
\fancypagestyle{plain}{%
  \fancyhf{}
  \fancyfoot[L]{\small \myshortauthors \hspace{0.1em} \textit{Preprint submitted to Elsevier}}
  \fancyfoot[R]{\small Page \thepage\ of \pageref{LastPage}}

}
\def\tsc#1{\csdef{#1}{\textsc{\lowercase{#1}}\xspace}}
\tsc{WGM}
\tsc{QE}
\tsc{EP}
\tsc{PMS}
\tsc{BEC}
\tsc{DE}

\begin{document}
\let\WriteBookmarks\relax
\shorttitle{Dynamic Bidirectional Pattern Memory}
\shortauthors{A.H. Lazem and W.J. Teahan}

\title[mode=title]{Dynamic Bidirectional Pattern Memory:
A Production-Scale Empirical Characterisation of
Inference-Time Gating in Clinical NLP}

\tnotemark[1]
\tnotetext[1]{This work was supported by the
              Ministry of Higher Education and Scientific Research in
              Iraq and conducted on the Supercomputing Wales Falcon
              (SCWF00175) platform.}

\author[1,2]{Ali H. Lazem}[%
    type=editor,
    role=Corresponding Author,
    orcid=0000-0002-8677-9689]
\cormark[1]
\ead{lhl23prg@bangor.ac.uk}
\credit{Conceptualization,
        Methodology,
        Software,
        Validation,
        Formal analysis,
        Investigation,
        Data curation,
        Writing -- original draft,
        Writing -- review \& editing,
        Visualization}

\author[1]{William Teahan}[%
    role=Co-Author,
    orcid=0000-0003-3640-6750]
\ead{w.j.teahan@bangor.ac.uk}
\credit{Conceptualization,
        Methodology,
        Supervision,
        Writing -- review \& editing,
        Project administration}

\affiliation[1]{organization={School of Computer Science and Engineering,
                              Bangor University},
                addressline={Bangor, Gwynedd},
                postcode={LL57 2DG},
                country={United Kingdom}}

\affiliation[2]{organization={University of Thi-Qar},
                addressline={Nasiriyah},
                postcode={64001},
                country={Iraq}}

\cortext[cor1]{Corresponding author}

\begin{abstract}
We study inference-time pattern-memory gating in a production-scale clinical natural
language processing (NLP) pipeline. The pipeline pairs a generator (Llama-3.3 70B)
that proposes extractions with a verifier (MMed-Llama-3.1 70B) that accepts or
rejects them, over 167{,}034 PMC-Patients narratives, and we add a lightweight memory
that learns at deployment time which extractions to filter, so the verifier need not
re-examine candidates already seen to fail. We report four findings. First, learning
these filtering rules directly from the verifier's rejections did not work at full
scale: the relation-extraction filter ended up empty even though the pipeline logged
785{,}797 rejected candidates, because the rejections were spread too thinly across
too many distinct forms to accumulate. Second, a simpler rule based on a fixed
clinical ontology produced the same filtering without relying on the verifier at all,
capturing 49{,}734 ontology-violating relations on a held-out 5{,}000-patient set.
Third, we tested five versions of the question-answering filter; four failed for
distinct and instructive reasons, and the fifth succeeded by checking whether a
patient's extracted clinical entities actually support the question being asked; on
the categories where it applies, it was 1.84 times more likely to flag an answer the
verifier would reject than one it would accept. Fourth, across all five versions one
pattern held: a filter is selective only when it tests the same evidence the verifier
itself weighs, not when it tries to imitate the verifier's output. Taken together,
these findings give a practical, transferable result for any generator-verifier
pipeline: the most natural memory design can fail silently at deployment scale, and
whether a pre-generation gate is selective is decided before any engineering effort,
by whether its signal probes the question the verifier itself answers. Throughout,
the system flags suspect extractions rather than deleting them, so every decision
stays visible for clinical review. All code and test artefacts are released openly.
\end{abstract}


\maketitle
\thispagestyle{plain}

\section{Introduction}
\label{sec:intro}
Large language models have transformed clinical natural language processing,
enabling extraction of named entities, relations, and question--answer pairs
from clinical narratives at scales unattainable with prior rule-based or
fine-tuned-classifier pipelines. A growing body of work pairs a generator model
with a verifier model that scores or accepts the generator's outputs
\citep{madaan2023selfrefine, shinn2023reflexion, yao2023react}, adding an
inference-time correctness check on top of instruction tuning and retrieval.
This pattern is attractive in clinical settings, where an unverified extraction
can carry a factual error into a patient record. At the scale of clinical
informatics corpora, however, where hundreds of thousands of narratives are
processed without model retraining, inference-time verification becomes the
dominant computational expense. This motivates a mechanism that avoids invoking
the verifier on candidates the pipeline has, in effect, already seen fail: an
inference-time memory that accumulates verdict evidence as the pipeline runs and
gates future candidates accordingly, without retraining either model.

The design space for such a memory is large, and which choices actually yield
\emph{selective} gating, gating that flags the candidates a verifier would
reject while leaving those it would accept, is not established. A memory can be
configured and yet remain empirically inactive at deployment scale; a gating
signal can be sophisticated and yet fail to separate the candidates the verifier
distinguishes. The central open question is therefore not how to build such a
memory, but which signal sources make its gates selective, and which do not.

This paper addresses that question through the design, deployment, and empirical
characterisation of Dynamic Bidirectional Pattern Memory (DBPM), an
inference-time pattern memory embedded in a multi-agent clinical-NLP pipeline.
The pipeline processes the 167{,}034-patient PMC-Patients corpus
\citep{zhao2023large} with Llama-3.3 70B as generator and MMed-Llama-3.1 70B
\citep{qiu2024mmed} as verifier. DBPM consumes the verifier's verdict stream and
gates the generator's future candidates at three pipeline stages (named-entity
recognition, relation extraction, and question answering with reasoning),
modifying no parameters of either model. We characterise its behaviour at
production scale, where failure modes become visible that benchmark-scale
evaluation does not expose.

The characterisation yields a single structural regularity that organises the
paper:a pre-generation gating signal is selective for verifier
rejection only when it directly tests the same evidence the verifier itself
weighs. A signal that probes a different quantity, however sophisticated its
aggregation, does not separate the candidates the verifier distinguishes. This
regularity is practical: a practitioner can assess a candidate signal source by
asking whether it probes the same question the verifier answers, before
committing engineering effort to it. It emerges from a controlled dissection of
five gating-signal sources, four of which fail to produce selective gating and
one of which succeeds, and it is consistent with a second, independent finding
about how such a memory populates at scale.

\subsection*{Contributions}
This paper makes four contributions, unified by the question of what makes
inference-time pattern-memory gating selective in a production clinical pipeline.

\paragraph{(C1) Production-scale evidence that the natural verifier-fed design
fails.} The most natural design, accumulating the verifier's own rejections by
candidate type and gating on the accumulated evidence, does not populate at
167{,}034-patient scale: the relation-extraction blocklist remains empty despite
hundreds of thousands of logged rejection events. We show the cause is
structural, an interaction between evidence accumulation, time-decay, and
pruning under a rejection stream that is diffuse rather than concentrated, and
that it generalises to any inference-time memory accumulating statistics over
free-form LLM output.

\paragraph{(C2) A working verifier-independent signal source for the same
channel.} Replacing the diffuse verifier-fed signal with a deterministic
ontology-violation predicate, which maps free-form relation labels to a bounded
canonical space, populates the same downstream gate where the verifier-fed
pathway did not. The signal source, not the persistence mechanism, is what
determines whether the channel populates.

 \paragraph{(C3) A five-source empirical dissection of gating-signal
selectivity.} We test five distinct signal sources for the question-answering
gate. Four fail to produce selective gating, each for a structurally distinct
reason, aggregation-level mismatch, signal absence, counting asymmetry, and
cross-task orthogonality, and one succeeds. The dissection isolates why each
failure occurs, rather than reporting only the design that worked.

\paragraph{(C4) A structural regularity for selective inference-time gating.}
The successful signal source is the one that tests whether a patient's extracted
clinical entities support the question being asked, the same grounding question
the verifier implicitly resolves. Across the five sources, selectivity tracks
this alignment and nothing else, yielding the regularity above as an empirical,
operationalisable guide for gate design in any generator--verifier architecture.

We deploy and measure DBPM under a measurement-instrumented discipline that flags
suspect candidates rather than deleting them, so every gating decision remains
visible for clinical review, and we disclose the operating-envelope bounds within
which the selective gate applies. All gating code and evaluation artefacts are
released openly.

\subsection*{Paper organisation}
Section~\ref{sec:related} positions DBPM against related work.
Section~\ref{sec:background} formalises the multi-agent pipeline and the
inference-time-memory problem, and defines the candidate signature on which the
memory is keyed. Section~\ref{sec:design} specifies the DBPM design, including
the early design decisions that motivated it, the persistence layer, three-tier
gating, the ontology-violation signal source, and the five gating-signal sources
evaluated for the question-answering gate. Section~\ref{sec:setup} describes the
experimental setup and the stratified 5{,}000-patient evaluation cohort.
Section~\ref{sec:results} reports the empirical findings, from the
verifier-fed-channel failure at production scale through the signal-source
dissection and its per-category operating envelope.
Section~\ref{sec:discussion} synthesises the structural regularity and its
clinical implications; Section~\ref{sec:limits} discloses the operating-envelope
and construct-overlap bounds; and Section~\ref{sec:future} identifies schema
closure, dataset-quality evaluation, and held-out validation as future work.
Section~\ref{sec:conclusion} concludes.

\section{Related Work}
\label{sec:related}
DBPM sits at the intersection of seven research threads in modern
NLP and machine learning. We position it against the closest
neighbour in each, drawing distinctions in mechanism, deployment
scope, and what the framework explicitly does {\it not} do.

\subsection{Multi-Agent Pipelines and Verifier-Driven Generation}
\label{sec:related-multiagent}

The generator–verifier separation underpins much recent inference-time work on LLM reliability. Self-Refine \citep{madaan2023selfrefine}
uses the same LLM to generate, critique, and refine outputs across
multiple self-iterations on a single input. Reflexion \citep{shinn2023reflexion} extends this with verbal self-reflection persisted in an
episodic memory buffer between trials, reinforcing the agent
through linguistic feedback rather than gradient updates. ReAct
\citep{yao2023react} interleaves reasoning traces with tool use
actions in a single trajectory. AutoGen \citep{wu2024autogen} and
MetaGPT \citep{hong2023metagpt} frame the verifier generator
separation as multi-agent conversation with role specialization
and standardized operating procedures, respectively. Generative
Agents \citep{park2023generative} extends the memory-trace concept to
long-running interactive simulations. LLM-as-a-judge frameworks
\citep{zheng2023judging} systematize the use of strong LLMs as
evaluators of weaker generators.

These systems share with DBPM the principle that an inference-time
critic can improve output quality without retraining. They differ
from DBPM in two structural ways. First, with the exception of Reflexion's
episodic buffer, the critic state in these systems either does not
persist across inputs (Self-Refine, ReAct) or persists as natural
language reflections that the generator must re-interpret each
call (Reflexion). DBPM persists categorical evidence about candidate
signatures in a structured store consulted by the gate at decision
time, with amortised $\mathcal{O}(1)$ lookup: gate queries reduce to hash-set
membership over precomputed per-task block, downgrade, and whitelist indices, and
writes to in-place dictionary updates keyed on the signature. This is required as
design constraint~C1 (Section~\ref{sec:background-memory-problem}) and detailed in
Appendix~\ref{app:implementation}. Second, none of these systems publishes
empirical evidence that the natural verifier-fed-memory design
fails at production scale; DBPM does (Section~\ref{sec:results-f1}), motivating the
verifier independent signal sources that follow.

\subsection{Knowledge Sharing Between Agents}
\label{sec:related-knowledge-sharing}

A line of work in multi-agent systems formalised structured
information exchange between cooperating agents long before the
LLM era. Tambe's STEAM model \citep{tambe1997flexible} framed
teamwork as agents maintaining shared joint intentions, monitoring
team-member performance, and selectively communicating evidence
under decision theoretic constraints an early operationalisation
of the principle that cooperating agents need shared structured
state, not only message-passing, to coordinate reliably.
\citet{teahan2006framework} subsequently proposed a framework for
knowledge sharing between autonomous agents operating in a shared
environment, addressing how candidate evidence should be exchanged
and consolidated.

DBPM operationalises the same problem for the generator–verifier
setting: the generator and the verifier are two agents that share
a structured memory consulted at gate time, with the verifier's
verdict statistics supplemented by a verifier independent
ontology signal forming the shared representation. The question
these earlier frameworks raised, ``How should autonomous agents
share evidence about candidate decisions?'', is the same question
DBPM answers in the modern setting. We view this lineage as a
relevant predecessor to the present paper's framing; the
contribution of our work is the empirical characterisation of
which signal sources actually populate such a shared memory at
deployment scale.

\subsection{Retrieval and Memory-Augmented Generation}
\label{sec:related-rag}

Retrieval-augmented generation \citep{lewis2020rag} and the
trillion-token RETRO architecture \citep{borgeaud2022retro} maintain
a deployment-time store consulted at inference, but the store
contains information (documents, embeddings, factual chunks)
used to augment the generator's input. End-to-end Memory Networks
\citep{sukhbaatar2015memnet}, Memory-Augmented Neural Networks
\citep{santoro2016mann}, and the Differentiable Neural Computer
\citep{graves2016dnc} introduce learned external memories used as
read/write substrate during training; the store evolves through
gradient updates and persists across episodes within a single
trained model.

The closest published cousin to DBPM in this lineage is Agentic
Context Engineering \citep{zhang2026ace}. ACE treats
context as an evolving playbook accumulated through a generator–reflector–curator loop, with structured incremental updates that
preserve detailed strategy knowledge across many interactions. ACE
and DBPM share the principle of inference-time-evolving deployment
state that informs future generations without weight modification.
They differ in mechanism layer: ACE updates a context document
that the agent reads (a prompt-side intervention); DBPM updates a
gate that filters the agent's output (a verification-side
intervention). Both operate downstream of a frozen generator;
neither requires retraining. We see ACE as the closest concurrent
work in evolving deployment state; DBPM addresses a different
question (``Which signals make a verification-gate selective?'') than
ACE (``How to grow a structured context?'').

\subsection{Continual Learning and Test-Time Adaptation}
\label{sec:related-continual}

A separate line of work studies how trained models can adapt without
forgetting prior knowledge. Elastic Weight Consolidation \citep{kirkpatrick2017ewc} selectively slows learning on weights important for prior tasks to mitigate catastrophic forgetting. Gradient Episodic
Memory \citep{lopezpaz2017gem} stores past examples and
constrains updates to avoid increasing loss on stored data. TENT
\citep{wang2021tent} adapts batch-normalization affine parameters
at test time via entropy minimization. SHOT \citep{liang2020shot}
freezes the source classifier and learns target features through
information maximization plus pseudo-labels. \citet{parisi2019continual}
review the broader continual-learning landscape.

DBPM is orthogonal to all of these. EWC, GEM, TENT, and SHOT all
modify model parameters either selectively (EWC, GEM) or via
self-supervised test-time objectives during deployment (TENT, SHOT). DBPM modifies
no parameters of either the generator or the verifier; it adds an
external gate consulted at inference. We cite this cluster to
position DBPM in the design space, not as direct competitor.

\subsection{Clinical NLP and Biomedical Extraction}
\label{sec:related-clinical-nlp}

End-to-end clinical NLP pipelines have evolved from rule-based
systems through neural and now generative architectures. MetaMap
\citep{aronson2001metamap} and cTAKES \citep{savova2010ctakes} established the modern pipeline structure with deterministic ontology mapping over
UMLS. Domain-specific BERTs, BioBERT \citep{lee2020biobert},
PubMedBERT \citep{gu2021pubmedbert} and ClinicalBERT \citep{alsentzer2019clinicalbert}---improved entity recognition through pretraining on biomedical corpora. The 2018 n2c2 shared task on adverse drug events \citep{henry2020n2c2} crystallized the relation-classification benchmark for clinical extraction. The verifier model in our pipeline, MMed-Llama-3.1 70B \citep{qiu2024mmed}, continues this trajectory with
multilingual medical pretraining.

Most recent clinical NLP systems emphasize the generator side
(model size, instruction tuning, domain pretraining) and treat the
pipeline as feed-forward: extraction stages produce candidates that
are scored or filtered downstream. DBPM adds an orthogonal
component, a deployment-time gate, that maintains structured
evidence about candidate signatures across patients and consults
this evidence at each stage. The ontology-based mechanism in our pipeline (the Ontology-Violation Filter; M3-RE in the codebase) explicitly returns to the deterministic-ontology tradition
of MetaMap and cTAKES, but applies it as a runtime gate alongside
LLM-driven extraction rather than as the primary extraction method.

\subsection{Inference-Time Adaptation and Weak Supervision}
\label{sec:related-inference-time}

Methods that adapt LLMs at inference without gradient updates
include few-shot in context learning \citep{brown2020gpt3}, chain-of-thought prompting \citep{wei2022cot}, and self-consistency
\citep{wang2023selfconsistency}. These operate within a single forward pass or a small number of self-iterations on a single input. They are
complementary to DBPM: a DBPM-gated pipeline can use any of these
at the generator stage. Weak-supervision frameworks \citep{ratner2019weaksup} combine multiple noisy label sources at training time without ground truth; DBPM combines multiple signal sources (verifier
verdicts, ontology violations) at deployment time. Self-training
methods such as Pseudo-Label \citep{lee2013pseudolabel} and Noisy Student \citep{xie2020noisystudent} use model predictions as training signals; DBPM does not train, but its memory store does accumulate verdict statistics that
could in principle feed downstream training (we do not pursue this
in the present paper).

\subsection{Knowledge Distillation and Data Curation}
\label{sec:related-distillation}

DBPM's output verified, gated extraction results is a candidate
substrate for downstream model training. Knowledge distillation \citep{hinton2015distillation} compresses an ensemble's predictions into a
single model. Stanford Alpaca \citep{taori2023alpaca} and Vicuna \citep{chiang2023vicuna} demonstrated that instruction tuning data
from strong generators can produce competitive smaller models.
LIMA \citep{zhou2023lima} showed that surprisingly little high
quality alignment data suffices for strong instruction following.
QuRating \citep{wettig2024qurating} systematizes quality-based selection
of pretraining data via LLM judgments on writing style, expertise,
and educational value.

DBPM is positioned relative to this cluster as an upstream
quality-control layer for data-generation pipelines. The framework
does not perform distillation or selection itself; it provides
structured provenance for downstream curators to use. The
dataset quality evaluation of DBPM's outputs (true-positive rates
on individual suppressions, downstream training-utility studies)
is outside the scope of this paper and is reserved for future work
on the resulting corpus.
\\

\textbf{Summary of positioning:} DBPM is not the first system to maintain inference-time state
(RAG, RETRO, Reflexion, ACE), not the first to use a verifier
distinct from a generator (Self-Refine, AutoGen, LLM-as-judge), and
not the first to apply ontology constraints in clinical extraction
(MetaMap, cTAKES). Its contribution is empirical: a characterization
of which signal sources actually populate inference-time pattern
memory at production scale, what failure modes prevent the naïve
verifier-fed design from working, and what structural property a
working signal source must satisfy. The five-iteration M1 dissection
(Section~\ref{sec:results-iterations}) and the Ontology-Violation Filter verifier-independent pathway
(Section~\ref{sec:results-ontology}) jointly answer the question of what makes inference-time pattern-memory gating selective in this regime. Table~\ref{tab:positioning} summarises this positioning against the closest system in each thread.

\begin{table}[t]
\centering
\caption{Positioning of DBPM against the closest system in each related thread.
A filled circle (\yes) marks a property the system has; an open circle (\no) marks
its absence. Columns are: inference-time state that persists across inputs; a verifier
distinct from the generator; a deterministic ontology constraint; published evidence
of a production-scale population failure; and treatment of which signal makes a gate
selective. DBPM is distinguished not by any single column but by the last two: it is
the only entry documenting the verifier-fed failure at deployment scale and isolating
the signal property that governs selectivity.}

\label{tab:positioning}
\footnotesize
\renewcommand{\arraystretch}{1.3}
\begin{tabular}{@{} l c c c c c @{}}
\toprule
& \textbf{Persistent} & \textbf{Verifier} & \textbf{Ontology} & \textbf{Scale-fail.} & \textbf{Signal} \\
\textbf{System} & \textbf{state} & \textbf{distinct} & \textbf{constraint} & \textbf{evidence} & \textbf{selectivity} \\
\midrule
Self-Refine \citep{madaan2023selfrefine}            & \no  & \yes & \no  & \no  & \no \\
Reflexion \citep{shinn2023reflexion}                & \yes$^{\dagger}$ & \yes & \no & \no & \no \\
ReAct \citep{yao2023react}                          & \no  & \no  & \no  & \no  & \no \\
RAG / RETRO \citep{lewis2020rag,borgeaud2022retro}  & \yes & \no  & \no  & \no  & \no \\
ACE \citep{zhang2026ace}                            & \yes & \yes & \no  & \no  & \no \\
MetaMap / cTAKES \citep{aronson2001metamap,savova2010ctakes} & \no & \no & \yes & \no & \no \\
LLM-as-judge \citep{zheng2023judging}               & \no  & \yes & \no  & \no  & \no \\
\midrule
\textbf{DBPM (this work)}                            & \yes & \yes & \yes & \yes & \yes \\
\bottomrule
\end{tabular}

\vspace{2pt}
{\footnotesize $^{\dagger}$ Reflexion persists natural-language reflections in an
episodic buffer, re-interpreted by the generator each call, rather than structured
categorical evidence consulted by a gate.}
\end{table}

\section{Background and Problem Formulation}
\label{sec:background}

This section formalises the deployment setting in which DBPM operates. The
objective is to specify the multi-agent pipeline, the verdict feedback signal it
produces, and the constraints any inference-time memory mechanism must satisfy.
Mechanism details are deferred to Section~\ref{sec:design}.

\subsection{Multi-Agent Clinical NLP Pipeline}
\label{sec:background-pipeline}

Let $D = \{x_1, \dots, x_N\}$ denote a corpus of $N$ clinical narratives, where
each $x_i$ is a single patient narrative. The pipeline produces structured
outputs across a finite task set $T$ with cardinality $|T| = 11$, indexed by a
task identifier $t \in T$. In our deployment, $N = 167{,}034$, the PMC-Patients
narrative corpus \citep{zhao2023large}. The eleven task buckets per patient
record are named entity recognition (NER), relation extraction (RE),
question--answer pairs with four-layer reasoning schemas (QAR), clinical-fidelity
summarisation, medication extraction, temporal-event extraction, active-risk
identification, risk-state-machine derivation, risk-based recommendations,
risk-grounded QA, and a visualisation-payload assembly stage. Each task emits its
own output stream: NER, RE, and QAR share the enriched multi-task output file,
while the other eight buckets each produce a dedicated \texttt{.jsonl} output.

DBPM's active consumer gates are defined for three of these tasks: NER, RE, and
QAR. This reflects the deployed worker's gate implementation; the remaining
buckets are not gated at the candidate level. NER is the upstream entity
extractor, and its output feeds downstream relation extraction, temporal-event
extraction, and the deterministic intelligence-layer components that depend on
extracted entities. RE and QAR carry their own gates because they emit
candidate-level verdict streams that DBPM consumes alongside NER. Throughout the
rest of this paper, all DBPM-gating claims refer to these three gates.

Each narrative is processed by two LLM agents:

\begin{itemize}
  \item A \textbf{generator agent} $G$ that produces, for each narrative-task
  pair $(x_i, t)$, a set of task-specific candidate outputs
  $C_{i,t} = \{c_{i,t}^{(k)} : k = 1, \dots, K_{i,t}\}$, where $c_{i,t}^{(k)}$ is
  the $k$-th candidate and $K_{i,t}$ is the number of candidates emitted for that
  pair. The generator is Llama-3.3 70B, served via vLLM continuous batching on
  two tensor-parallel H200 GPUs.

  \item A \textbf{verifier agent} $V$ that assigns each candidate $c_{i,t}^{(k)}$
  a categorical verdict $v_{i,t}^{(k)} \in V_t$, where $V_t$ denotes the
  verdict space for task $t$. (The verdict space $V_t$ is a set of categorical
  labels and is notationally distinct from the verifier agent $V$, which is a
  model; the two share a letter but not a referent.) The verdict space is
  task-specific: $V_t = \{\mathrm{PASS}, \mathrm{FAIL}\}$ for the binary
  verification tasks (NER, QAR, medications, summary), and
  $V_t = \{\mathrm{PASS\_STRONG}, \mathrm{PASS\_WEAK}, \mathrm{FAIL}\}$ for
  relation extraction. The graded form on RE reflects clinical relations of
  varying support strength. The remaining non-gated tasks use verdict structures
  that DBPM does not consume and that we do not specify here. The verifier emits
  no continuous confidence; the pipeline derives any post-hoc multiplier from the
  categorical verdict and internal rule-based signals. The verifier is
  MMed-Llama-3.1 70B \citep{qiu2024mmed}, served on the remaining two H200 GPUs.
\end{itemize}

The QAR task produces question--answer pairs together with a four-layer reasoning
schema. A single DBPM gate (denoted \texttt{gate\_qa} in the codebase, for
historical reasons) handles both the minimal-context QA case and the
full-reasoning QAR case. The deployed runs reported throughout this paper use the
QAR configuration, and all selectivity claims
(Sections~\ref{sec:results-iterations} and~\ref{sec:results-envelope}) are
measured against QAR outputs.

Each candidate $c_{i,t}^{(k)}$ admits an extractable \textbf{signature}
$\sigma(c_{i,t}^{(k)}) \in \Sigma_t$. We define a signature as the canonical key
under which candidates are grouped in memory: a deterministic, normalised
identifier computed from the candidate, such that two candidates occurring in
different narratives are treated as the same pattern if and only if they share a
signature. Here $\Sigma_t$ is the task-specific signature space, the set of all
signatures admissible for task $t$, and $\Sigma = \bigsqcup_{t \in T} \Sigma_t$
is their disjoint union across tasks. The signature is the unit of memory: the
memory stores and retrieves evidence keyed on $\sigma$, not on the raw candidate.
Throughout, a \emph{signature} is the unique key under which evidence is grouped,
and a \emph{pattern} is the stored record (severity, count, timestamp) that this
evidence accumulates into; many events may share one signature and therefore update
one pattern, but a signature identifies at most one pattern per task.
Across the three gated tasks, the signatures are defined as follows.

\begin{itemize}
  \item NER signatures are the lowercase-normalised entity string.
  \item RE signatures are the uppercase-normalised triple
  $(\mathrm{h\_type}, r, \mathrm{t\_type})$ of head category, relation label $r$,
  and tail category. For RE we additionally retain the raw head and tail entity
  strings as auxiliary fields, used only by the NER-coverage gate
  (Section~\ref{sec:design-qar}); these auxiliary fields are not part of the RE
  signature.
  \item QAR signatures are the lowercase-normalised question template plus an
  error-class disambiguator, introduced in the first iteration of the QAR gate
  (Section~\ref{sec:design-qar}).
\end{itemize}

\subsection{The Verdict Stream}
\label{sec:background-verdict-stream}

As the pipeline processes the documents in $D$, it emits a stream of verdict
events:
\begin{equation}
  S = \{(\sigma_j, t_j, v_j, \tau_j, s_j) : j = 1, \dots, M\}
  \label{eq:verdict-stream}
\end{equation}

\noindent where, for the $j$-th event, $\sigma_j \in \Sigma_{t_j}$ is the
candidate signature, $t_j \in T$ is the task that produced the event,
$v_j \in V_{t_j}$ is the verdict the verifier (or a rule-based source) assigned,
$\tau_j$ is the wall-clock timestamp at which the event was recorded, and $s_j$
is the signal source label identifying the subsystem that produced the event
(for example, the direct verifier, a rule-based ontology check, or a cross-task
propagation path). The source label $s_j$ selects the source weight $w_s$ applied
during the memory update (Section~\ref{sec:design}). The total number of events
in the stream is $M$. Each rejection event is persisted to
\texttt{universal\_rejections.jsonl} as the tuple
$(\mathrm{uid}, \mathrm{stage}, \mathrm{entity}, \mathrm{reason}, \mathrm{conf})$:
the patient identifier, the pipeline stage at which the rejection occurred, the
rejected entity or candidate, a categorical rejection reason, and the associated
confidence. The exact event counts by stage and reason at production scale are
reported in Section~\ref{sec:results-f1} (Table~\ref{tab:rejection-composition}).

In a static pipeline, each verdict $v_j$ is consumed once its candidate is
accepted or rejected, and the stream $S$ is then discarded. The central design
question for DBPM is whether $S$ instead contains exploitable structure. Two
questions follow. First, does the verdict distribution conditioned on a signature
concentrate as evidence accumulates? Second, can a memory that estimates this
distribution online convert it into gating decisions for future candidates with
the same signature? Section~\ref{sec:results} addresses both empirically across
several signal sources; the answer differs by source.

\subsection{The Inference-Time Memory Problem}
\label{sec:background-memory-problem}

A solution to the inference-time pattern-memory problem maintains, after the
$j$-th event, a memory state that is queried at gate time. We require the state
to yield, for any signature-task pair $(\sigma, t)$, a quantitative estimate of
how strongly accumulated evidence opposes a candidate and, optionally, how
strongly it supports one. We capture this with two real-valued functions, indexed
by the event count $j$:

\begin{itemize}
  \item $B_j : \Sigma \times T \to [0, 1]$, the \textbf{blocklist severity}: the
  value $B_j(\sigma, t)$ is the severity accumulated against signature $\sigma$ on
  task $t$ after $j$ events, with higher values indicating stronger evidence for
  rejection.
  \item $W_j : \Sigma \times T \to [0, 1]$, the \textbf{whitelist confidence}:
  the value $W_j(\sigma, t)$ is the confidence accumulated for signature $\sigma$
  on task $t$ after $j$ events, with higher values indicating stronger evidence
  for acceptance.
\end{itemize}

The memory begins with no prior beliefs, $B_0(\sigma, t) = W_0(\sigma, t) = 0$
for all $(\sigma, t)$, and is updated after each event by an update operator $U$:
\begin{equation}
  (B_j, W_j) = U(B_{j-1}, W_{j-1}, \sigma_j, t_j, v_j, \tau_j, s_j)
  \label{eq:update-operator}
\end{equation}

\noindent where $(B_{j-1}, W_{j-1})$ is the memory state before the event,
$(\sigma_j, t_j, v_j, \tau_j, s_j)$ is the $j$-th verdict event defined in
Equation~\eqref{eq:verdict-stream}, and $(B_j, W_j)$ is the updated state. The
choice of $U$, together with the gating policy that consumes $(B_j, W_j)$, must
satisfy four practical constraints imposed by the deployment setting.

\begin{description}
  \item[(C1) Constant-time gating.] Querying $B_j(\sigma, t)$ and $W_j(\sigma, t)$
  must complete in $\mathcal{O}(1)$ amortised time. Gates are queried at every
  candidate-generation event and cannot become a pipeline bottleneck at
  production throughput.
  \item[(C2) Bounded memory size.] The signature space $|\Sigma|$ is unbounded in
  principle, since NER signatures are arbitrary normalised strings; the number of
  stored signatures must nonetheless remain tractable, and the storage policy
  must admit eviction.
  \item[(C3) Bounded values.] $B_j(\sigma, t), W_j(\sigma, t) \in [0, 1]$ for all
  $(\sigma, t)$ and all $j$. Boundedness prevents a single high-frequency
  signature from dominating gating over long deployment horizons.
  \item[(C4) Stability under non-stationarity.] Clinical documentation
  conventions evolve across specialties, time periods, and institutional sources,
  so the memory must admit recovery: evidence must attenuate once it ceases to be
  reinforced. As we note here and substantiate in Section~\ref{sec:results-f1},
  the choice of decay parameters interacts with the distributional shape of the
  rejection stream. When rejections are diffuse across many signatures, the decay
  window can exhaust an individual signature's severity before it accumulates to
  threshold. This interaction is the central empirical finding of
  Section~\ref{sec:results-f1}.
\end{description}

Together, (C1)--(C4) carve out the design space. The remaining choices, how $U$
aggregates evidence, how it forgets, which signal sources contribute, and how the
gating policy converts $(B_j, W_j)$ into a decision, are the subject of
Section~\ref{sec:design}.

\subsection{Notation Summary}
\label{sec:background-notation}

Table~\ref{tab:notation} summarises the notation used throughout
Sections~\ref{sec:background} and~\ref{sec:design}. Each symbol is also defined in
the text at the point of first use; the table is provided as a consolidated
reference.

\begin{table}[h]
  \centering
  \caption{Notation used throughout Sections~\ref{sec:background}
  and~\ref{sec:design}. Additional notation (learning rates $\eta_{v,t}$, source
  weights $w_s$, decay parameters $\rho_v$ and $h_t$, gating thresholds
  $\theta_t^{\mathrm{block}}$ and $\theta_t^{\mathrm{down}}$) is introduced at the
  points it is needed in Section~\ref{sec:design}.}
  \label{tab:notation}
  \renewcommand{\arraystretch}{1.2}
  \begin{tabular}{ll}
    \hline
    \textbf{Symbol} & \textbf{Meaning} \\
    \hline
    $x_i$, $D$, $N$ & Narrative, corpus, corpus size ($N = 167{,}034$) \\
    $T$, $|T|$ & Task set; $|T| = 11$ in this deployment \\
    $t \in T$ & Task index \\
    $G$, $V$ & Generator (Llama-3.3 70B), verifier (MMed-Llama-3.1 70B) \\
    $c_{i,t}^{(k)}$, $C_{i,t}$, $K_{i,t}$ & Candidate, candidate set, count per $(i, t)$ \\
    $\sigma \in \Sigma_t$, $\Sigma$ & Signature, per-task and disjoint-union signature space \\
    $v_{i,t}^{(k)}$, $V_t$ & Verdict, verdict space for task $t$ \\
    $\tau_j$ & Wall-clock timestamp of the $j$-th event \\
    $s_j$ & Signal source label of the $j$-th event (selects weight $w_s$, \S\ref{sec:design}) \\
    $S$, $M$ & Verdict event stream, total event count \\
    $B_j(\sigma, t)$, $W_j(\sigma, t)$ & Blocklist severity, whitelist confidence \\
    $U$ & Memory update operator (\S\ref{sec:design}) \\
    $\sigma_0(v)$ & Initial severity for a new signature (\S\ref{sec:design-params}) \\
    \hline
  \end{tabular}
\end{table}

\section{DBPM Design}
\label{sec:design}


\begin{figure*}[!htbp]
\centering
\begin{tikzpicture}[
    font = \sffamily,
    llm/.style = {rectangle, rounded corners=2pt,
                  draw=black!75, line width=0.5pt,
                  fill=black!6,
                  minimum width=45mm, minimum height=15mm,
                  align=center, font=\sffamily\small},
    dbpm/.style = {rectangle, rounded corners=4pt,
                   draw=black!85, line width=1.0pt,
                   fill=teal!12,
                   minimum width=76mm, minimum height=56mm,
                   align=center, font=\sffamily\small},
    gatebox/.style = {rectangle, rounded corners=2pt,
                      draw=black!60, line width=0.4pt,
                      fill=white,
                      minimum width=68mm, minimum height=8mm,
                      align=center, font=\sffamily\scriptsize,
                      inner sep=2pt},
    storebox/.style = {rectangle, rounded corners=2pt,
                      draw=black!60, line width=0.4pt,
                      fill=white,
                      minimum width=32mm, minimum height=12mm,
                      align=center, font=\sffamily\scriptsize,
                      inner sep=2pt},
    ontobox/.style = {rectangle, rounded corners=2pt,
                      draw=black!60, line width=0.4pt,
                      fill=orange!15,
                      minimum width=68mm, minimum height=8mm,
                      align=center, font=\sffamily\scriptsize,
                      inner sep=2pt},
    flow/.style    = {-{Stealth[length=2.6mm, width=2.0mm]},
                      line width=0.7pt, draw=black!80},
    feedback/.style = {-{Stealth[length=2.6mm, width=2.0mm]},
                       line width=0.6pt, draw=black!65,
                       dashed, dash pattern=on 2pt off 1.5pt},
    elabel/.style = {font=\sffamily\scriptsize, fill=white,
                     inner sep=3pt, align=center},
]

\node[llm] (G) at (0, 5.5)
    {\textbf{Generator} \textit{G}\\[1pt]
     \scriptsize Llama-3.3\,70B\\[-1pt]
     \scriptsize (vLLM, TP=2)};

\node[dbpm] (DBPM) at (0, 0) {};

\node[llm] (V) at (0, -5.5)
    {\textbf{Verifier} \textit{V}\\[1pt]
     \scriptsize MMed-Llama-3.1\,70B\\[-1pt]
     \scriptsize (vLLM, TP=2)};

\node[align=center, font=\sffamily\small]
    at ($(DBPM.north) + (0, -5mm)$)
    {\textbf{DBPM Gate}\\[-1pt]
     \scriptsize gated tasks:\,NER,\,RE,\,QAR};

\node[gatebox] at ($(DBPM.center) + (0, 14mm)$) (GateFn)
    {gate function: BLOCK / DOWNGRADE / ALLOW};

\node[storebox] at ($(DBPM.center) + (-18mm, -1mm)$) (Block)
    {\textbf{Blocklist} $B_j$\\
     severity per $\sigma$};
\node[storebox] at ($(DBPM.center) + (18mm, -1mm)$) (White)
    {\textbf{Whitelist} $W_j$\\
     confidence per $\sigma$};

\node[ontobox] at ($(DBPM.center) + (0, -16mm)$) (Onto)
    {Ontology-Violation Filter (verifier-independent)};

\draw[flow] (G.south) -- (DBPM.north);
\node[elabel, right=3mm] at ($(G.south)!0.5!(DBPM.north)$) {
    \makebox[0pt][l]{candidates $C_{i,t}$}\\
    \makebox[0pt][l]{signatures $\sigma$}
};

\draw[flow] (DBPM.south) -- (V.north);
\node[elabel, right=3mm, align=left] at ($(DBPM.south)!0.5!(V.north)$) {
    BLOCK $\Rightarrow$ dropped\\
    DOWNGRADE $\Rightarrow$ tagged\\
    ALLOW $\Rightarrow$ passes
};

\draw[feedback] (V.east) -- (5.0, -5.5) -- (5.0, 0) -- (DBPM.east);

\node[font=\sffamily\scriptsize, right=2mm, align=left] at (5.0, -2.75) {
    \textbf{Feedback Loop:}\\
    verdict stream $\{(\sigma, t, v, \tau, s)\}$\\
    \textit{(no gradient updates)}
};

\node[font=\sffamily\scriptsize\itshape, text=black!55, left, align=right] at (-5.2, 5.5)
    {generator zone};
\node[font=\sffamily\scriptsize\itshape, text=black!55, left, align=right] at (-5.2, 0)
    {inference-time\\pattern memory};
\node[font=\sffamily\scriptsize\itshape, text=black!55, left, align=right] at (-5.2, -5.5)
    {verifier zone};

\end{tikzpicture}

\caption{Architecture and locus of DBPM. The pattern memory sits between the
generator ($G$, Llama-3.3 70B) and the verifier ($V$, MMed-Llama-3.1 70B). It
gates candidate outputs before $V$ is called (forward path, solid arrows) and
accumulates evidence from $V$'s verdicts after each call (feedback loop, dashed
arrow). The gate returns BLOCK, DOWNGRADE, or ALLOW for each candidate signature
$\sigma$ (Equation~\ref{eq:gate}): a BLOCK'ed candidate is dropped before $V$ is
called, a DOWNGRADE'ed candidate passes through with a tag in its output
metadata, and an ALLOW'ed candidate is unchanged. The dual store is the heart of
the persistence layer (Section~\ref{sec:design-severity}): the blocklist $B_j$
accumulates severity evidence, while the whitelist $W_j$ records direct verifier
confirmation and takes precedence over indirect evidence. The Ontology-Violation
Filter (Section~\ref{sec:design-ontology}) provides a verifier-independent signal
source feeding the same blocklist. DBPM updates no parameters of $G$ or $V$.}
\label{fig:architecture}
\end{figure*}
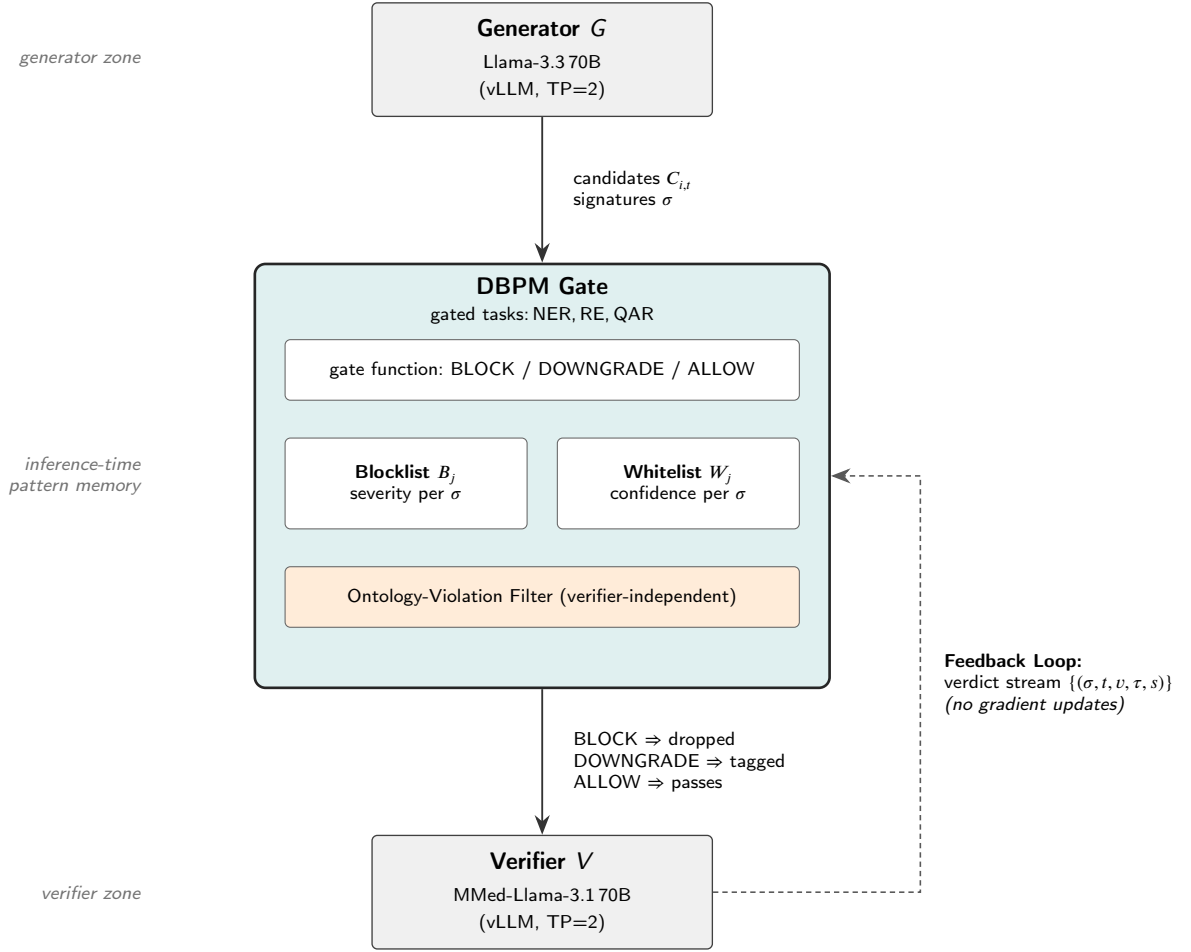

This section specifies the design of DBPM. We organise it around four questions a
reader can hold in mind throughout: how a single verdict updates the memory
(Section~\ref{sec:design-severity}), how evidence about one task informs another
(Section~\ref{sec:design-crosstask}), how the accumulated evidence becomes a
gating decision (Section~\ref{sec:design-gating}), and how the memory forgets so
that it tracks changing clinical language (Section~\ref{sec:design-decay}). Two
further components follow: a verifier-independent signal source that populates the
relation-extraction channel where the verifier-fed signal does not
(Section~\ref{sec:design-ontology}), and the question-answering gate, whose five
design iterations are the empirical core of the paper
(Section~\ref{sec:design-qar}). Figure~\ref{fig:architecture} shows where DBPM
sits between the generator and verifier, and
Algorithm~\ref{alg:dbpm-update} gives the update operator in full.

The parameter values quoted throughout are read directly from the deployed
worker. To keep the body readable, the in-code identifiers for flags, thresholds,
and methods are confined to Appendix~\ref{app:implementation}; the body names each
mechanism by what it does.

\subsection{Early Design Decisions}
\label{sec:design-early}

DBPM began from the most natural design for an inference-time memory: let the
verifier's own rejections accumulate by signature, and gate any future candidate
whose signature has accumulated enough rejection evidence. The intuition is
direct. If the verifier has rejected a given kind of candidate often enough, the
pipeline should stop paying to re-verify it and should instead flag it on sight.
This design has three moving parts, each independently reasonable: an additive
rule that raises a signature's severity on each rejection, a decay that lowers
severity over time so the memory can recover when clinical language shifts, and a
pruning step that discards signatures whose severity has fallen to negligible
levels.

We deployed this design first and expected it to populate the blocklist at
production scale. It did so for two of the three gated tasks but not for the
third, relation extraction, and the reason it failed there
(Section~\ref{sec:results-f1}) shaped every design choice that follows. The
verifier-independent signal source of Section~\ref{sec:design-ontology} exists
because the verifier-fed source did not populate the relation-extraction channel;
the five iterations of the question-answering gate
(Section~\ref{sec:design-qar}) exist because selectivity, not population, was the
binding constraint there. We present the mechanism as deployed, but the order in
which its parts were added was driven by these two empirical failures, and we
return to that narrative when the evidence for it is in hand
(Section~\ref{sec:results}).

\subsection{Parameters}
\label{sec:design-params}

The mechanism is governed by a small set of constants, all fixed at worker
initialisation and none tuned to the empirical results of
Section~\ref{sec:results}. We introduce each here so that the equations of the
following subsections can refer to it directly. Table~\ref{tab:learning-rates}
and Table~\ref{tab:source-weights} collect the two multi-valued sets; the
remaining scalars are stated inline.

\paragraph{Learning rates $\eta_{v,t}$.}
The learning rate $\eta_{v,t}$ sets how much a single verdict of type $v$ on task
$t$ raises a signature's severity (Equation~\ref{eq:severity-update}). The values
appear in Table~\ref{tab:learning-rates}. Relation extraction carries higher
rates than the other two tasks because its verifier emits a graded verdict
(\texttt{PASS\_STRONG}, \texttt{PASS\_WEAK}, \texttt{FAIL}) rather than a binary
one, so each relation verdict carries more information.

\begin{table}[h]
\centering
\caption{Per-task verdict learning rates $\eta_{v,t}$ for the three DBPM-gated
tasks. The values are fixed at worker initialisation and are not tuned to the
empirical results of Section~\ref{sec:results}.}
\label{tab:learning-rates}
\renewcommand{\arraystretch}{1.2}
\begin{tabular}{lccc}
\hline
\textbf{Verdict} & \textbf{NER} & \textbf{RE} & \textbf{QAR} \\
\hline
\texttt{hard\_fail}       & 0.15 & 0.20 & 0.15 \\
\texttt{soft\_downgrade}  & 0.05 & 0.08 & 0.05 \\
\texttt{success}          & 0.03 & 0.05 & 0.04 \\
\hline
\end{tabular}
\end{table}

\paragraph{Source weights $w_s$.}
Each verdict is scaled by a weight reflecting how much the subsystem that produced
it should be trusted. A direct verifier verdict carries full weight ($1.0$); a
rule-based ontology check carries half ($0.5$); cross-task signals, which are
indirect by construction, carry less ($0.2$ to $0.4$, by direction). The full set
is in Table~\ref{tab:source-weights}. One source, a generator self-critique
weighted $0.7$, is reserved in the code but inactive in every run reported here.

\paragraph{Decay and recovery.}
Two distinct decay mechanisms operate on different clocks. An in-run
multiplicative factor $\rho_v$ damps repeated firing of the same signature within
a run ($\rho_{\mathrm{hard\_fail}} = 0.99$,
$\rho_{\mathrm{soft\_downgrade}} = 0.985$). A wall-clock half-life $h_t$ governs
how fast an un-reinforced signature fades across days ($h_{\mathrm{NER}} = 5$,
$h_{\mathrm{RE}} = h_{\mathrm{QAR}} = 10$ days). The two are kept separate
deliberately: one tracks the event sequence, the other elapsed time. A third
constant $\rho_{\mathrm{success}} = 0.998$ exists but is never reached on the
severity path, since success events update the whitelist and return before the
severity update.

\paragraph{Initial severity $\sigma_0(v)$.}
A first-time rejection initialises severity to $\sigma_0(\mathrm{hard\_fail}) =
0.50$ or $\sigma_0(\mathrm{soft\_downgrade}) = 0.25$, scaled by the source weight.

\paragraph{Gating thresholds.}
A signature is blocked once its severity exceeds the per-task block threshold
$\theta_t^{\mathrm{block}}$ ($0.65$ NER, $0.70$ RE, $0.60$ QAR) and it has been
seen at least $n_t^{\min}$ times ($3$, $2$, $3$ respectively). It is downgraded,
rather than blocked, when severity lies in the band between the downgrade
threshold $\theta_t^{\mathrm{down}} = 0.40$ (all tasks) and the block threshold,
with a single observation sufficient.

\paragraph{Storage caps $C_t$.}
At most $C_t$ signatures are retained per task ($200$ NER, $300$ RE, $150$ QAR);
when the count is exceeded, the lowest-severity entries are evicted. The caps were
set so the working set fits in process memory at the deployed throughput.

\subsection{Severity Update}
\label{sec:design-severity}

For each signature $\sigma$ on task $t$, the blocklist stores its current severity
$\sigma_{\mathrm{sev}} \in [0,1]$, an observation count $c$, the timestamp
$\tau_{\mathrm{last}}$ of its most recent update, and a cross-task hit counter
$\kappa$ used in Section~\ref{sec:design-crosstask}. The whitelist stores a
confidence $\sigma_{\mathrm{conf}} \in [0,1]$ in place of severity.

A verdict event is routed by its verdict type. A \emph{success} takes the
whitelist path, raising the signature's confidence toward a ceiling of $0.99$
(initialised at $0.90$ on first sighting) and returning without touching severity.
A \emph{hard\_fail} or \emph{soft\_downgrade} takes the blocklist path. For a
signature seen for the first time, severity is initialised to the source-weighted
base $w_s \cdot \sigma_0(v)$. For a signature already in memory, the update
applies the in-run decay before adding the new evidence:
\begin{equation}
\sigma_{\mathrm{sev}} \leftarrow
   \min\bigl\{1,\; \rho_v \cdot \sigma_{\mathrm{sev}}
                 + w_s \cdot \eta_{v,t}\bigr\},
\qquad c \leftarrow c + 1,\;
       \tau_{\mathrm{last}} \leftarrow \tau.
\label{eq:severity-update}
\end{equation}

Equation~\eqref{eq:severity-update} carries the substance of the update rule, and
each term earns its place. The decay factor $\rho_v < 1$ stops a single signature
from saturating at the ceiling under repeated firing, which is what purely
additive updates would do. The additive term scales the base learning rate
$\eta_{v,t}$ by the source weight $w_s$, so an indirect signal moves severity less
than a direct verifier verdict. The outer clip to $[0,1]$ enforces boundedness
(constraint C3 of Section~\ref{sec:background-memory-problem}).

\subsection{Cross-Task Propagation}
\label{sec:design-crosstask}

A rejection on one task is often evidence about another. If the verifier rejects a
relation, the entities it joins are themselves suspect; if it rejects an entity,
no relation built on that entity can be sound. DBPM acts on this through
pattern-level cross-task propagation: a rejection on one task records derived
evidence on a related task's memory. Three directions are active, and their
asymmetry is deliberate.

When a relation is hard-failed, DBPM records a downgrade against each of its head
and tail entity strings in the NER memory, at the cross-task weight and with the
increment further halved, because the evidence is twice removed from a direct NER
verdict. A hard-failed question-answer pair propagates the same way to the
entities named in its answer text, at the lowest weight in the system, since
long-form answers mix genuine entities with prose. These two directions are
graded: they nudge severity upward without ever, on their own, blocking a
candidate. The third direction is categorical. When an entity is hard-failed at
the NER stage, it is added to a preemption set, and any future relation candidate
built on that entity is dropped before the verifier sees it. This direction is
all-or-nothing rather than graded because an entity the verifier has already
rejected cannot meaningfully take part in any relation, so a partial signal would
be wasted.

Four invariants keep propagation safe. \emph{Whitelist protection}: a signature
the verifier has directly confirmed receives no propagated evidence, so direct
confirmation always dominates indirect inference. \emph{Propagation cap}: once a
signature has received propagated evidence three times ($\kappa \geq 3$), further
propagation is suppressed, with $\kappa$ itself decaying so the suppression lifts
after sustained quiet. \emph{Severity-zone guard}: once a signature's severity
reaches $0.55$, cross-task evidence can no longer raise it; indirect evidence may
push a signature into the downgrade band but never into the block band on its own.
\emph{Recursion prevention}: a propagated update is itself ineligible to
propagate, which rules out cycles. Together the last two invariants give the
structural guarantee made precise in Section~\ref{sec:design-gating}: cross-task
evidence alone cannot block a candidate.

These pattern-level directions are distinct from the decision-time cross-task
signals used by the difficulty gate and the NER-coverage gate
(Section~\ref{sec:design-qar}), which read cross-task state without writing to it.
Pattern-level propagation is active in every run reported in
Section~\ref{sec:results}; its isolated effect is below the pipeline's noise floor
and is quantified in Appendix~\ref{app:crosstask}.

\subsection{Three-Tier Gating}
\label{sec:design-gating}

The accumulated evidence becomes a decision at three gate sites, one per gated
task. For each candidate, the gate returns one of three verdicts:
\begin{equation}
\mathrm{gate}_j(\sigma, t) =
\begin{cases}
\mathrm{ALLOW}     & \text{if } \sigma \in W_j^t \\
\mathrm{BLOCK}     & \text{if } c(\sigma, t) \geq n_t^{\min}
                     \text{ and } B_j(\sigma, t) > \theta_t^{\mathrm{block}} \\
\mathrm{DOWNGRADE} & \text{if } c(\sigma, t) \geq 1
                     \text{ and } \theta_t^{\mathrm{down}} < B_j(\sigma, t) \leq \theta_t^{\mathrm{block}} \\
\mathrm{ALLOW}     & \text{otherwise.}
\end{cases}
\label{eq:gate}
\end{equation}

The decision is read top to bottom. A signature on the whitelist exits
immediately as ALLOW, never reaching the severity checks; this early exit is why
verifier-confirmed candidates are never downgraded
(Section~\ref{sec:results-downgrade}). Otherwise the signature is blocked if its
severity clears the block threshold and it has enough corroborating observations,
downgraded if its severity falls in the band below that threshold, and allowed by
default. A blocked candidate is dropped before the verifier is called. A
downgraded candidate proceeds, but its confidence is multiplied by $0.7$, a
reduction that flows through relation filtering and graph weighting without
removing the candidate. An allowed candidate is untouched.

Four properties of this policy are worth stating plainly. \emph{Whitelist
precedence} is unconditional: no amount of accumulated indirect evidence can
override a direct success. \emph{Minimum support} on the block clause means no
single event can block a signature; at least $n_t^{\min}$ observations must agree,
whereas a single observation suffices to flag a borderline candidate for
downgrade. \emph{Threshold separation} ($\theta_t^{\mathrm{down}} <
\theta_t^{\mathrm{block}}$) leaves a non-empty downgrade band on every task, so
the gate degrades gracefully rather than flipping between block and allow.
\emph{Direct-evidence requirement for block} follows from the severity-zone guard
of Section~\ref{sec:design-crosstask}: cross-task-only severity is capped at
$0.55$, below every task's block threshold of $0.60$ or more, so a candidate
reaches BLOCK only if its history includes at least one direct verifier or
rule-based verdict. This is a structural separation, not a tuned heuristic.

\subsection{Decay and Pruning}
\label{sec:design-decay}

The in-run decay of Equation~\eqref{eq:severity-update} handles repeated firing
within a run. A second mechanism handles the passage of time. At each save, every
blocklist severity is multiplied by a wall-clock decay term:
\begin{equation}
\sigma_{\mathrm{sev}} \leftarrow \sigma_{\mathrm{sev}}
   \cdot 2^{-\Delta\tau / h_t},
\qquad \Delta\tau = \tau_{\mathrm{now}} - \tau_{\mathrm{last}}
\text{ (days).}
\label{eq:wallclock-decay}
\end{equation}
A signature that stops being reinforced loses half its severity every $h_t$ days,
so the gate follows shifts in clinical documentation over a long deployment.
Equation~\eqref{eq:wallclock-decay} is the operational form of the recovery
constraint (C4 of Section~\ref{sec:background-memory-problem}). Once a severity
falls below $0.01$, the signature is pruned. The cross-task counter $\kappa$ also
decays at each prune, so a signature that once hit the propagation cap is not
quarantined permanently. A per-task storage cap then evicts the lowest-severity
survivors if the count exceeds $C_t$.

This is where the natural design meets its binding condition, and the outcome turns
on an interaction between mechanisms rather than on any single part. Whether a
channel populates depends on a race between reinforcement and decay. If a signature
is reinforced faster than it decays, it clears threshold; if it is reinforced more
slowly, it is pruned before it accumulates. Section~\ref{sec:results-f1} shows that
at production scale the relation-extraction channel loses this race, and that the
cause is the distributional \emph{shape} of the rejection stream rather than its
volume.

\subsection{The Ontology-Violation Filter}
\label{sec:design-ontology}

Section~\ref{sec:results-f1} establishes that the verifier-fed relation-extraction
channel produces no usable patterns at production scale. The fix is not to change
the memory but to change the signal that feeds it. The Ontology-Violation Filter (hereafter ``the Filter")
is a deterministic check that fires on every relation candidate, independent of
the verifier, flagging any relation whose head-type, relation, tail-type triple is
not admissible under the worker's clinical ontology:
\begin{equation}
\begin{aligned}
&\mathrm{ontology\_violation}(h_{\mathrm{type}}, r, t_{\mathrm{type}}) \;\leftrightarrow\; \\
&\quad (h_{\mathrm{type}}, t_{\mathrm{type}})
   \in \mathrm{PRIORS} \;\;\textbf{and}\;\;
   \mathrm{alias}(r) \notin \mathrm{PRIORS}\bigl[(h_{\mathrm{type}}, t_{\mathrm{type}})\bigr],
\end{aligned}
\label{eq:ontology}
\end{equation}
where $\mathrm{PRIORS}$ is a static table of $29$ admissible type-triples encoding
the ontology, and $\mathrm{alias}(\cdot)$ is a $32$-entry map that canonicalises
verbal variants of a relation (for example, \texttt{MAY\_REVEAL} and
\texttt{REVEALS} collapse to one label). Both are fixed at load time, and the
comparison is case-normalised.

A flagged candidate is recorded as a hard failure keyed on its type-triple, with
the rule-based source weight, and persisted through the same decay and pruning
machinery as any other blocklist entry. The Ontology-Violation Filter introduces no new gate: it
populates the same relation-extraction blocklist that the existing relation gate
already queries, so the verifier-fed pathway and the Ontology-Violation Filter feed one channel from
two independent sources. The reason it succeeds where the verifier-fed pathway
fails is the reason developed in Section~\ref{sec:design-decay}. Because the
predicate fires on a fixed alphabet of type-triples rather than on the open
vocabulary of natural-language relation labels, its evidence concentrates on a
handful of signatures and wins the reinforcement-versus-decay race that the
diffuse verifier-fed signal loses.

\subsection{The Question-Answering Gate: Five Iterations}
\label{sec:design-qar}

The question-answering gate is the most heavily worked component of DBPM and the
empirical heart of the paper. Here the binding problem was never whether the
channel would populate, it did, but whether the gate would be \emph{selective}:
whether it would flag the candidates the verifier rejects while leaving the rest
alone. We tested five signal sources in turn. Four were not selective, each
failing for a structurally different reason; the fifth was. We present them in the
order we tried them, because the sequence is itself the argument: the failures are
not a catalogue of mistakes but a map of which kinds of signal can possibly make
such a gate selective.

The gate fires once per question-answer pair, keyed on the pair's QAR category
(one of eleven QAR categories, comprising seven general clinical categories,
\texttt{diagnosis}, \texttt{treatment}, \texttt{symptoms}, \texttt{tests},
\texttt{history}, \texttt{observation}, and \texttt{risk\_factor}, and four
outcome-related categories, \texttt{outcome\_clinicalstatus},
\texttt{outcome\_disposition}, \texttt{outcome\_mortality}, and
\texttt{adverse\_event}). The five sources
differ only in what they read to make that decision.

\paragraph{First: category-aggregate fail-mass (within-task).}
The first source totals, per category, the accumulated rejection mass and flags
categories above a threshold. It is anti-selective: the categories it flags are
the high-volume ones, not the error-prone ones, and the verifier does not decide
category by category. The failure mode is an \emph{aggregation-level mismatch}: the
gate decides at the category level while the verifier decides at the pair level.

\paragraph{Second: verifier uncertainty (within-task).}
The second source reads a per-pair field the verifier sets when its text output
fails to parse, on the hypothesis that this marks uncertainty. The field is almost
never set: modern instruction-tuned models reliably produce parseable output, so
the signal is effectively empty at scale. The failure mode is \emph{signal absence}:
the hoped-for signal does not exist in the data.

\paragraph{Third: per-category fail-rate (within-task).}
The third source estimates each category's true fail-rate by dividing its rejection
count by its total:
\begin{equation}
\hat{P}(\mathrm{FAIL} \mid \mathrm{cat}) =
   \frac{\mathrm{count}_{\mathrm{FAIL}}[\mathrm{cat}]}
        {\mathrm{count}_{\mathrm{FAIL}}[\mathrm{cat}] +
         \mathrm{count}_{\mathrm{SUCCESS}}[\mathrm{cat}]},
\label{eq:m1v5-failrate}
\end{equation}
firing a downgrade above a category fail-rate threshold. The fail-rate varies
across categories, so the signal is not degenerate, yet the gate is still
anti-selective. The cause is subtle: failures are counted per pair while successes
are counted per question template (roughly eleven templates total), so the
denominator and numerator are aggregated at different units and the estimate
overstates the true rate for high-volume categories. The failure mode is a
\emph{counting asymmetry}: numerator and denominator are tallied at mismatched
granularities.

\paragraph{Fourth: per-patient difficulty (cross-task).}
The fourth source is the first to read cross-task state at decision time. It
combines the category fail-rate with a per-patient difficulty score built from
upstream rejection density:
\begin{equation}
\mathrm{difficulty}(\mathrm{patient}) =
   \frac{\mathrm{rejections\_upstream}(\mathrm{patient})}
        {\max\bigl(\mathrm{rejections\_upstream}(\mathrm{patient})
          + \mathrm{ner\_yield}(\mathrm{patient}),\; 1\bigr)},
\label{eq:m1v6-difficulty}
\end{equation}
where $\mathrm{rejections\_upstream}$ counts the patient's NER- and RE-stage
verifier rejections and $\mathrm{ner\_yield}$ counts its verifier-passed entities.
The gate fires a downgrade only when both the category fail-rate and the patient
difficulty exceed their thresholds. It produces no measurable selectivity: the
difficulty distributions for verifier-passed and verifier-failed pairs are
statistically indistinguishable. The failure mode is \emph{cross-task orthogonality}:
upstream extraction difficulty is unrelated to what the verifier judges at the
answer stage, which turns on answer content such as hallucination or
over-specification.

\paragraph{Fifth: NER-coverage intersection (cross-task).}
The fifth source asks a different question, the one the verifier itself implicitly
answers: does the patient have the extracted clinical entities that an answer in
this category would need to be grounded? It checks whether the QAR category being
asked about appears among the patient's accepted NER categories:
\[
\mathrm{ner\_cats}(\mathrm{patient}) =
   \{\,\mathrm{fact.category}\,:\,
     \mathrm{fact} \in \mathrm{accepted\_facts}(\mathrm{patient})\,\},
\]
and gates accordingly:
\begin{equation}
\mathrm{gate}(\mathrm{cat}, \mathrm{patient}) =
\begin{cases}
\mathrm{ALLOW}     & \text{if } \mathrm{cat} \in
                     \mathrm{ner\_cats}(\mathrm{patient}) \\
\mathrm{DOWNGRADE} & \text{if } \mathrm{cat} \notin
                     \mathrm{ner\_cats}(\mathrm{patient})
                     \text{ and the category has a matching NER bucket} \\
\text{fall back to the fourth source} & \text{otherwise.}
\end{cases}
\label{eq:m1v7-gate}
\end{equation}
A patient with matching coverage passes; a patient without it is flagged. Nine of
the eleven categories have a matching NER bucket; the two that do not
(\texttt{adverse\_event}, \texttt{outcome\_mortality}) fall back to the fourth
source.

This source was motivated by a read-only test on the development cohort: across
qualified categories, a patient lacking the relevant NER coverage was
substantially more likely to have its answer rejected than a patient with it. We
measure the deployed gate's selectivity by its \emph{lift}, the ratio of its
flagging rate on verifier-rejected pairs to its flagging rate on verifier-accepted
pairs:
\begin{equation}
\mathrm{lift} =
  \frac{P(\mathrm{gate\ flags} \mid \mathrm{verifier} = \mathrm{FAIL})}
       {P(\mathrm{gate\ flags} \mid \mathrm{verifier} = \mathrm{PASS})},
\label{eq:lift}
\end{equation}
where flagging means a DOWNGRADE or BLOCK verdict. A lift above one means the gate
is more likely to flag a pair the verifier rejects than one it accepts.
Section~\ref{sec:results-iterations} reports this lift on both cohorts, along with
the per-category operating envelope and a structural-correctness check; the fifth
source is the working design, and the principle that explains why it works and the
other four fail is the subject of Section~\ref{sec:discussion}.

\subsection{DOWNGRADE-First Measurement Discipline}
\label{sec:design-downgrade}

Throughout the measurements of this paper, the gate runs in a
measurement-instrumented mode that tags suspect candidates rather than removing
them. A class-level flag controls whether the policy emits BLOCK verdicts at all;
when it is off, the BLOCK tier of Equation~\eqref{eq:gate} is suppressed and a
candidate that would otherwise be blocked is instead tagged in its per-pair output
without being removed from the pipeline stream. This is the default in every
measurement reported here. The gate therefore records which candidates \emph{would}
be blocked under a hard-blocking configuration, without committing to the removal.

The discipline has a direct empirical consequence, developed in
Section~\ref{sec:results-downgrade}: because the NER-coverage gate tags rather than
removes, its ON-versus-OFF ablation arm comparisons produce near-zero pair-count deltas even
when the gate tags thousands of pairs. Selectivity must therefore be read at the
gate-tagging level, through the lift ratio of Equation~\eqref{eq:lift}, not from
output-volume change. A production deployment that wishes to act on the gate's
verdicts would enable the hard-blocking configuration, converting each tag into a
BLOCK; the in-code flag for this is given in Appendix~\ref{app:implementation}.

\begin{algorithm}[t]
\caption{DBPM update operator $U$ (per verdict event)}
\label{alg:dbpm-update}
\begin{algorithmic}[1]
\Require verdict event $(\sigma, t, v, \tau, s)$;
         memory state $(B_{j-1}, W_{j-1})$;
         recursion guard \textit{is\_cross\_task} (default \textsc{False})
\Ensure  updated memory state $(B_j, W_j)$
\Statex
\If{$v = \textsc{success}$}
    \State update $W_{j-1}(\sigma, t)$ via the whitelist path
        \Comment{whitelist path, Sec.~\ref{sec:design-severity}}
    \State \Return $(B_{j-1}, W_j)$
\EndIf
\Statex
\State $\eta \gets \eta_{v,t}$; \quad $w \gets w_s$
    \Comment{learning rate, source weight}
\If{entry $e$ for $(\sigma, t)$ exists in $B_{j-1}$}
    \State $e.\mathrm{sev} \gets \min\!\big(1,\; \rho_v\, e.\mathrm{sev} + w\,\eta\big)$
        \Comment{severity update, Sec.~\ref{sec:design-severity}}
    \State $e.\mathrm{count} \gets e.\mathrm{count} + 1$; \quad $e.\tau \gets \tau$
\Else
    \State initialise $e$ with $\mathrm{sev} = w\,\sigma_0(v)$, $\mathrm{count}=1$, $\kappa=0$, $\tau=\tau$
    \State insert $e$ into $B_j$
\EndIf
\Statex
\If{\textit{is\_cross\_task} \textbf{or} $v \neq \textsc{hard\_fail}$}
    \State \Return $(B_j, W_{j-1})$
        \Comment{guards rule out cycles, Sec.~\ref{sec:design-crosstask}}
\EndIf
\Statex
\If{$t = \textsc{RE}$}
    \For{$e_{\mathrm{str}} \in \{\mathrm{head}, \mathrm{tail}\}$ with $4 < |e_{\mathrm{str}}| < 60$}
        \State $\Call{Propagate}{e_{\mathrm{str}}, w_{\mathrm{cross\_task\_re}}, \tau}$
            \Comment{RE\,$\to$\,NER, Sec.~\ref{sec:design-crosstask}}
    \EndFor
\ElsIf{$t = \textsc{QAR}$}
    \For{$e_{\mathrm{str}} \in \mathrm{split}(\mathrm{answer}, \texttt{","})$}
        \State $\Call{Propagate}{e_{\mathrm{str}}, w_{\mathrm{cross\_task\_qa}}, \tau}$
            \Comment{QAR\,$\to$\,NER, Sec.~\ref{sec:design-crosstask}}
    \EndFor
\ElsIf{$t = \textsc{NER}$}
    \State $P \gets P \cup \{\sigma\}$
        \Comment{NER\,$\to$\,RE preemption set, Sec.~\ref{sec:design-crosstask}}
\EndIf
\Statex
\State \Return $(B_j, W_{j-1})$
\Statex
\Procedure{Propagate}{$e_{\mathrm{str}}, w, \tau$}
    \Comment{graded cross-task update, Sec.~\ref{sec:design-crosstask}}
    \If{$e_{\mathrm{str}} \in W_{j-1}^{\textsc{NER}}$ \textbf{or} $\kappa(e_{\mathrm{str}}) \geq 3$
        \textbf{or} $B_j(e_{\mathrm{str}}, \textsc{NER}) \geq 0.55$}
        \State \textbf{return}
            \Comment{whitelist, cap, and severity-zone guards}
    \EndIf
    \State $s_{\mathrm{before}} \gets B_j(e_{\mathrm{str}}, \textsc{NER})$
    \State recurse: $U(e_{\mathrm{str}}, \textsc{NER}, \textsc{soft\_downgrade}, \tau, w)$
           with \textit{is\_cross\_task} $=$ \textsc{True}
    \State $\Delta \gets B_j(e_{\mathrm{str}}, \textsc{NER}) - s_{\mathrm{before}}$
    \State $B_j(e_{\mathrm{str}}, \textsc{NER}) \gets s_{\mathrm{before}} + \tfrac{1}{2}\Delta$
        \Comment{half-weight propagation}
    \State $\kappa(e_{\mathrm{str}}) \gets \kappa(e_{\mathrm{str}}) + 1$
\EndProcedure
\end{algorithmic}
\end{algorithm}

\begin{table}[h]
\centering
\caption{Source weights $w_s$ applied in the severity update
(Equation~\ref{eq:severity-update}). Each weight scales a verdict event by the
reliability of the subsystem that produced it: a direct verifier verdict carries
full weight, indirect and cross-task signals carry less. The self-critique source
is reserved in the code but inactive in all runs reported here.}
\label{tab:source-weights}
\small
\setlength{\tabcolsep}{8pt}
\renewcommand{\arraystretch}{1.2}
\begin{tabular}{llc}
\toprule
\textbf{Source} & \textbf{Signal origin} & \textbf{Weight $w_s$} \\
\midrule
\texttt{verifier\_mmed}   & Direct MMed-Llama verdict        & 1.0 \\
\texttt{rule\_based}      & Ontology / regex check           & 0.5 \\
\texttt{cross\_task\_ner} & NER-to-RE preemption             & 0.4 \\
\texttt{cross\_task\_re}  & RE-to-NER propagation            & 0.3 \\
\texttt{cross\_task\_qa}  & QAR-to-NER propagation           & 0.2 \\
\texttt{cross\_task}      & Generic cross-task fallback      & 0.5 \\
\midrule
\texttt{verifier\_self}   & Generator self-critique (inactive) & (0.7) \\
\bottomrule
\end{tabular}
\end{table}

\section{Experimental Setup}
\label{sec:setup}

This section specifies the dataset, the evaluation cohort, and the ablation
protocol used to evaluate DBPM. The design goal of the protocol is to isolate the
contribution of each DBPM component through controlled paired ablations on a single
locked 5{,}000-patient cohort, and to separately characterise the verifier-fed
channel at full 167{,}034-patient production scale. Hardware and software
configuration, the resume protocol, and the in-code flag mappings are given in
Appendix~\ref{app:implementation}; a single
consolidated summary of every fixed setting appears in Table~\ref{tab:config}.

\subsection{Dataset}
\label{sec:setup-dataset}

Evaluation uses the multi-task derivative of the PMC-Patients corpus that the
pipeline produces. The source corpus is PMC-Patients~\citep{zhao2023large}, a
publicly released collection of $N = 167{,}034$ patient case reports drawn from
PubMed Central full-text articles. The pipeline loads and processes the corpus
end to end, converting each narrative into a structured record carrying the eleven
task-bucket outputs of Section~\ref{sec:background-pipeline}, of which three (NER,
RE, QAR) are DBPM-gated and feed the analysis of this paper.

\paragraph{The held-out 5{,}000-patient cohort.}
The three paired ablations of Section~\ref{sec:setup-ablation} share a single
5{,}000-patient subsample, drawn once from the full corpus by stratified random
sampling and then fixed across every ablation arm so that ON-versus-OFF comparisons
are patient-aligned. The cohort is stratified along three dimensions: age bucket
(0--17, 18--39, 40--64, 65+, plus an explicit unknown bucket for missing or
unparseable ages), gender (F, M, unknown), and document-length tertile computed
from the full corpus, with cut points T1 $\leq$ 1{,}931 characters $<$ T2 $\leq$
3{,}121 characters $<$ T3. The three-dimensional grid yields 24 non-empty strata.

The deployed pipeline processed the entire 167{,}034-patient corpus end to end over
approximately twelve days; Table~\ref{tab:production-run} characterises this full
production run, the aggregate yields, throughput, and precision against which the
ablation cohort below is drawn. Of these narratives, 163{,}400 are human case
reports and 3{,}634 are veterinary; we retain both, since the gating mechanism
operates on extraction structure rather than on any species-specific content, but
note that the corpus is not exclusively human.

Sampling proceeds in two stages. Let $n_{\mathrm{target}} = 5{,}000$ be the target
cohort size and $n_{\mathrm{strata}} = 24$ the number of non-empty strata. Each
non-empty stratum first receives a minimum floor allocation of
$\min(10,\, \max(1,\, n_{\mathrm{target}} / (2\, n_{\mathrm{strata}})))$ patients,
capped by the stratum's population, so that rare strata remain visible. The
remaining sample size is then distributed proportionally to each stratum's residual
population above its floor, with rounding remainders assigned to the
largest-deficit strata so the total reaches exactly $n_{\mathrm{target}} = 5{,}000$. Selection within a
stratum is uniform random without replacement, seeded for reproducibility, and the
final identifier list is shuffled to randomise processing order.

The resulting cohort mirrors the demographic skew of the source corpus: the five
largest strata all fall in the 40--64 age bucket and together contribute roughly a
third of the cohort, with the 18--39 female bucket dominating the next tier.
Because the same cohort is reused across all arms, this skew is held constant and
does not confound the paired comparisons. The full stratum allocations, tertile cut
points, and identifier list are released as \texttt{ablation\_5k\_uids.json}.

Every gated-task output is verifier-attested: candidates that fail verification are
logged to \texttt{universal\_rejections\allowbreak.jsonl} with stage and reason metadata, so
the dataset records the complete candidate-and-verdict stream that drives DBPM. At
production scale, this stream contains 1{,}367{,}615 events; the per-stage breakdown
(Table~\ref{tab:rejection-composition}) and its interaction with the persistence
layer are reported in Section~\ref{sec:results-f1}.

\begin{table}[t]
\centering
\caption{Full-run production characterisation on the 167{,}034-patient corpus, with
all values aggregated from the deployed run's per-patient metrics and final memory
state. The run spanned approximately twelve days (4--16 May 2026). The relation rejection count (785{,}797) and the empty relation blocklist are
analysed in Section~\ref{sec:results-f1}; the high relation precision (0.994)
reflects the permissive-verifier regime discussed in
Section~\ref{sec:limits-precision}.
}
\label{tab:production-run}
\footnotesize
\renewcommand{\arraystretch}{1.1}
\begin{tabular}{@{} l r @{}}
\toprule
\textbf{Quantity} & \textbf{Value} \\
\midrule
\multicolumn{2}{@{}l}{\textit{Corpus and throughput}} \\
Patient narratives processed            & 167{,}034 \\
\quad human / veterinary                & 163{,}400 / 3{,}634 \\
Elapsed wall-clock duration             & {\raise.17ex\hbox{$\scriptstyle\sim$}}12 days \\
Mean end-to-end time per patient        & 102.9\,s \\
\quad NER / RE / QAR stage means        & 24.6 / 14.7 / 43.1\,s \\
\addlinespace[0.3em]
\multicolumn{2}{@{}l}{\textit{Extraction yield (corpus totals)}} \\
Named entities                          & 4{,}471{,}110 \\
Candidate relations                     & 4{,}915{,}334 \\
Relations accepted to enriched output   & 4{,}129{,}537 \\
Question--answer pairs                  & 1{,}035{,}199 \\
\addlinespace[0.3em]
\multicolumn{2}{@{}l}{\textit{Verifier rejections and persistence}} \\
Relation rejections (classifier stage)  & 785{,}797 \\
Persisted relation blocklist patterns   & 0 \\
Persisted NER / QA patterns             & 5{,}860 / 11 \\
\addlinespace[0.3em]
\multicolumn{2}{@{}l}{\textit{Precision (corpus means)}} \\
Relation precision                      & 0.994 \\
Verifier precision                      & 0.945 \\
Final-output precision                  & 0.888 \\
Mean entity coverage                    & 0.773 \\
\bottomrule
\end{tabular}
\end{table}

\subsection{Ablation Protocol}
\label{sec:setup-ablation}

We evaluate DBPM through four configurations, summarised in
Table~\ref{tab:ablation-protocol}. The first three are paired ablations on the
held-out 5{,}000-patient cohort, each isolating one component by toggling a single
flag while holding the rest of the system fixed. The fourth is a production-scale
cross-check on the full corpus, not a paired comparison: it exists to characterise
whether the verifier-fed relation-extraction channel populates at deployment scale,
which is a property only visible at full volume.

\begin{table}[h]
\centering
\caption{The four evaluation configurations. Rows 1--3 are paired ablations on the
held-out 5{,}000-patient cohort, each isolating one DBPM component. Row 4 is a
production-scale cross-check on the full 167{,}034-patient corpus, not a paired
comparison. Flag names are the literal switches toggled, given here for
reproducibility; their concept mapping is in Appendix~\ref{app:implementation}.}
\label{tab:ablation-protocol}
\renewcommand{\arraystretch}{1.25}
\small
\begin{tabular}{p{0.5cm}p{3.6cm}p{3.6cm}p{4.2cm}}
\hline
\textbf{\#} & \textbf{Configuration} & \textbf{What is tested} & \textbf{What is held constant} \\
\hline
1 & Ontology Filter ON vs OFF (\texttt{BPM\_DISABLE\_M3}) & The Ontology-Violation Filter (Section~\ref{sec:design-ontology}) & Held-out cohort; NER-coverage gate ON in both arms; pattern-level propagation active in both arms \\
2 & NER-coverage gate ON vs OFF (\texttt{M1V7\_ENABLE}) & The NER-coverage intersection gate (Section~\ref{sec:design-qar}) & Held-out cohort; Ontology Filter ON in both arms; difficulty gate as fallback in both arms \\
3 & DBPM full ON vs full OFF (\texttt{BPM\_DISABLE} master switch) & All new DBPM components simultaneously & Held-out cohort; legacy \texttt{\_ner\_block} / \texttt{\_re\_block} checks active in both arms \\
4 & 167K production cross-check & Population of the verifier-fed RE channel at deployment scale & Full 167{,}034-patient corpus; deployed DBPM-full configuration \\
\hline
\end{tabular}
\end{table}

Each paired ablation records three families of measurement per arm: task-level pair
counts (per-stage candidate counts and verifier pass/fail counts for NER, RE, and
QAR); DBPM-specific metrics (per-task BLOCK / DOWNGRADE / ALLOW counts,
pattern-level cross-task propagation count, and whitelist and blocklist sizes by
severity bucket); and system metrics (per-stage runtime, candidate-rejection rate
by stage, and end-to-end patient-processing time).

All four configurations run under the DOWNGRADE-first measurement discipline of
Section~\ref{sec:design-downgrade}: the gate tags candidates in the output without
removing them from the stream. Selectivity is therefore measured at the tagging
level, via the lift ratio of Equation~\eqref{eq:lift}, and provenance is measured by
the count of gate-tagging events relative to total pipeline pair throughput, rather
than by output-volume change.

One further component, pattern-level cross-task propagation
(Section~\ref{sec:design-crosstask}), is active in the deployed configuration but is
not among the four configurations above. Its isolated effect is measured by a
separate paired ablation (cross-task propagation ON vs OFF) reported in
Appendix~\ref{app:crosstask}. We keep it out of the main protocol because its
measured contribution, a pair-count delta of 0.098\% of pipeline throughput, falls
below the pipeline's per-stage non-determinism floor and so carries no main-results
weight.

\section{Results}
\label{sec:results}

This section reports what the data supports and discusses its limitations. The
paper is a methodological characterisation of DBPM as an inference-time
pattern-memory gating mechanism: which signal sources populate its channels, which
gate designs are selective and which fail, and what structural pattern emerges
across the design space. Dataset-quality evaluation of DBPM's outputs (human-rated
true-positive rates, downstream clinical-correctness scoring) is the subject of
separate forthcoming work on the resulting corpus and is excluded from this paper's
claims.

The findings rest on four empirical anchors: a production-scale run revealing the
structural absence of verifier-fed signal accumulation
(Section~\ref{sec:results-f1}); a paired ablation of the Ontology-Violation Filter
on the held-out cohort (Section~\ref{sec:results-ontology}); a dissection of the
five question-answering gate signal sources, with development-cohort replication
(Sections~\ref{sec:results-iterations} and~\ref{sec:results-envelope}); and a
confirmation that DBPM operates by tagging rather than suppression
(Section~\ref{sec:results-downgrade}).

\subsection{Production-Scale Failure of the Verifier-Fed Channel}
\label{sec:results-f1}

The most natural design, accumulating the verifier's own rejections by relation
signature and gating on the accumulated evidence, did not populate at production
scale. The deployed run processed the full 167{,}034-patient corpus over
approximately twelve days of elapsed wall-clock time (4--16 May 2026), spanning
several resubmitted SLURM allocations (Appendix~\ref{app:resume}), and its full
characterisation is given in Table~\ref{tab:production-run}. At the end of the run,
the verifier-fed relation-extraction blocklist held no patterns at all: the relation
channel was absent from the persisted memory state, not merely below its threshold of
count $\geq 2$ and severity $> 0.70$. The named-entity and question-answering
channels, by contrast, populated normally over the same run, holding 5{,}860 and 11
patterns respectively in the final state.

This empty channel is the more striking because the same run produced a large volume
of relation rejections. Of the 4{,}915{,}334 candidate relations the generator
emitted, 785{,}797 were rejected, logged as Stage-3 classifier events on the
relation-extraction pathway (Table~\ref{tab:rejection-composition}). The channel that
should have accumulated these rejections recorded none of them.

\paragraph{Two senses of ``rejection,'' and why the distinction is the finding.}
The reconciliation of 785{,}797 rejections with zero persisted patterns is not a
discrepancy; it is the mechanism. The 785{,}797 figure counts classifier-stage
rejection events on the candidate stream. These are distinct from memory-fail events,
the rejections that the persistence layer actually consolidates into a stored
signature. In the deployed run the relation memory's fail counter remained at zero:
no rejection was ever consolidated, so no signature was ever created, so the channel
stayed empty. The rejections existed in abundance on the stream; they simply never
became persistable memory.

The cause is the interaction of three persistence-layer mechanisms, each individually
sensible (Section~\ref{sec:design-decay}): per-signature aggregation under the
head-type, relation, tail-type key; wall-clock severity decay with a ten-day
half-life on relation patterns (Equation~\ref{eq:wallclock-decay}); and prune-on-save
below a severity floor of $0.01$. Because DBPM state persists across the resubmitted
allocations and the decay advances on real UTC time rather than compute time
(Appendix~\ref{app:resume}), the relevant clock is the run's full twelve-day span,
over which the ten-day half-life fires roughly once. Across that window, relation
rejections recurred too infrequently per signature relative to the decay: each
signature's severity decayed between save events faster than reinforcement arrived,
so signatures fell below the floor and were pruned before they could accumulate. A
signature reinforced rarely never survives long enough to clear threshold, no matter
how many rejections occur in total, because they are spread across too many distinct
signatures.

\begin{table}[t]
\centering
\caption{Rejection-stream composition at 167{,}034-patient production scale. Each
event is persisted to \texttt{universal\_rejections.jsonl} as the tuple
$(\mathrm{uid}, \mathrm{stage}, \mathrm{entity}, \mathrm{reason}, \mathrm{conf})$.
The Stage-3 classifier and Stage-4 QA verifier dominate; the Stage-1 gatekeeper
rejections are ontology-mismatch events across clinical categories. The 785{,}797
Stage-3 classifier rejections are the relation-pathway events that, despite their
volume, did not persist into the verifier-fed blocklist channel.}
\label{tab:rejection-composition}
\small
\setlength{\tabcolsep}{6pt}
\renewcommand{\arraystretch}{1.2}
\begin{tabular}{llr}
\toprule
\textbf{Stage} & \textbf{Rejection reason} & \textbf{Events} \\
\midrule
Stage 3 (Classifier)  & \texttt{SOTA\_Classify\_None}             & 785{,}797 \\
Stage 4 (QA verifier) & \texttt{QA\_Verifier\_Reject}             & 548{,}093 \\
\midrule
Stage 1 (Gatekeeper)  & \texttt{Ontology\_Mismatch\_Diagnosis}    &  20{,}246 \\
                      & \texttt{Ontology\_Mismatch\_Tests}        &  10{,}105 \\
                      & \texttt{Ontology\_Mismatch\_Risk\_Factor} &   2{,}061 \\
                      & \texttt{Ontology\_Mismatch\_History}      &   1{,}244 \\
                      & \texttt{Ontology\_Mismatch\_Observation}  &      21 \\
                      & \texttt{Ontology\_Mismatch\_Outcome}      &       1 \\
\midrule
Stage 1 (Verifier)    & \texttt{NER\_Verifier\_Reject}            &      33 \\
Stage 1 (Rescue)      & \texttt{Rescued\_By\_Confidence}          &      14 \\
\midrule
\textbf{Total}        &                                           & \textbf{1{,}367{,}615} \\
\bottomrule
\end{tabular}
\end{table}

\begin{figure}[H]
    \centering
    \includegraphics[width=\linewidth]{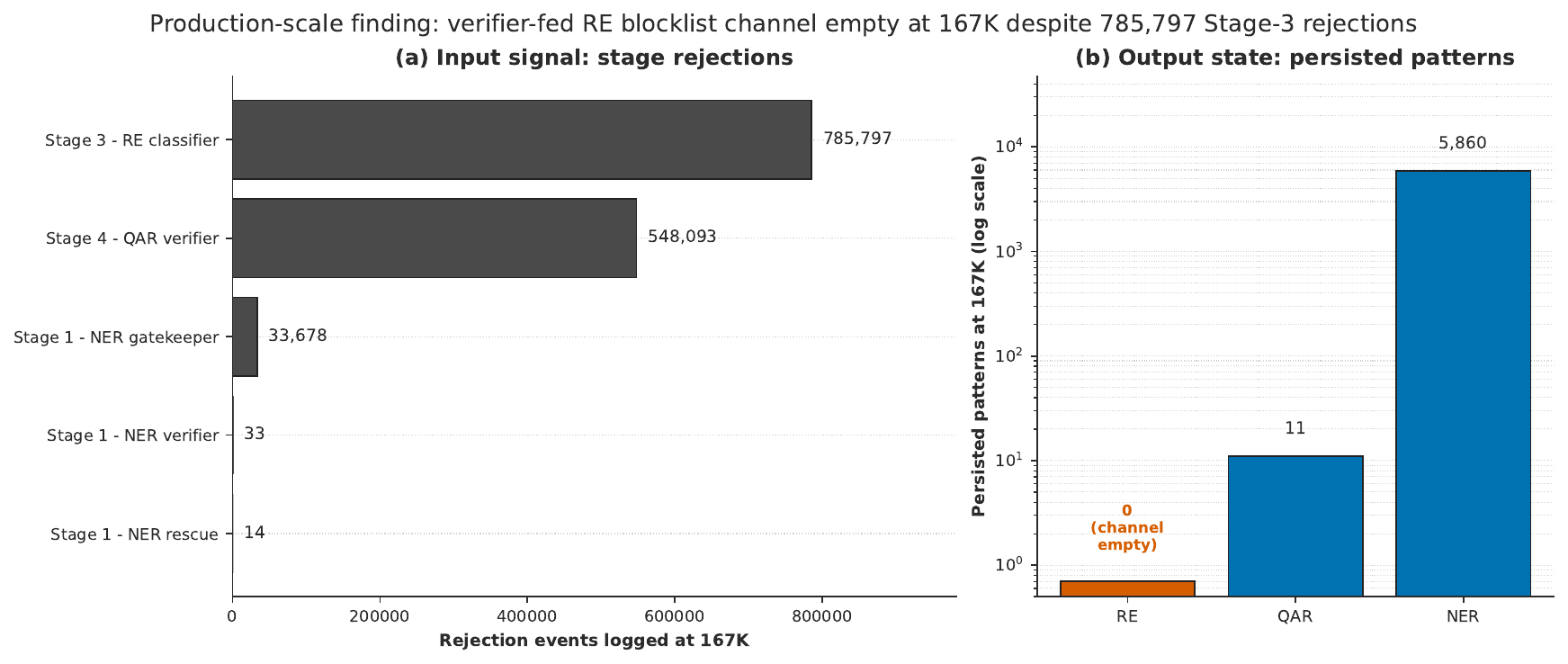}
    \caption{The verifier-fed relation-extraction persistence channel at
    167{,}034-patient deployment scale. \textbf{(a)} Input signal: 1.37M rejection
    events logged across the five verifier stages, dominated by Stage~3 (relation
    classifier, 785{,}797 events, reason \texttt{SOTA\_Classify\_None}) and Stage~4
    (QA verifier, 548{,}093 events). \textbf{(b)} Output state: persisted patterns
    at threshold in the relation, QA, and named-entity channels after the run (log
    scale). The relation channel is empty despite consuming 785{,}797 input events;
    the named-entity channel populates with 5{,}860 patterns and the QA channel with
    11. This asymmetry is the central production-scale finding and motivates the
    verifier-independent Ontology-Violation Filter (Section~\ref{sec:design-ontology}).}
    \label{fig:f1-channel}
\end{figure}

The other two gated channels received more concentrated per-signature traffic and
populated normally at the same final snapshot, the question-answering channel holding
11 records and the named-entity channel 5{,}860. The collapse is therefore specific to
the relation channel's rejection distribution, not a property of the persistence layer
in general: the identical decay-and-prune machinery that emptied the relation channel
left the other two intact. Figure~\ref{fig:f1-channel} contrasts the input volume
against the persisted output state.

The finding generalises beyond this pipeline. Any inference-time memory that
accumulates per-signature statistics over free-form LLM output is exposed to the same
diffusion: when the output vocabulary is open, rejections spread across too many
distinct signatures for any one to be reinforced faster than it decays, and the
channel stays empty regardless of total volume. The mitigation is to change the signal
source so that events concentrate on a bounded alphabet, which is exactly what the
Ontology-Violation Filter of Section~\ref{sec:design-ontology} (evaluated next) does.

\subsection{The Ontology-Violation Filter: A Verifier-Independent Source}
\label{sec:results-ontology}

We evaluate the Ontology-Violation Filter on the held-out cohort under a paired
ON/OFF ablation, with the NER-coverage gate held ON in both arms and the system in
DOWNGRADE-first mode.

\begin{figure}[t]
    \centering
    \includegraphics[width=\linewidth]{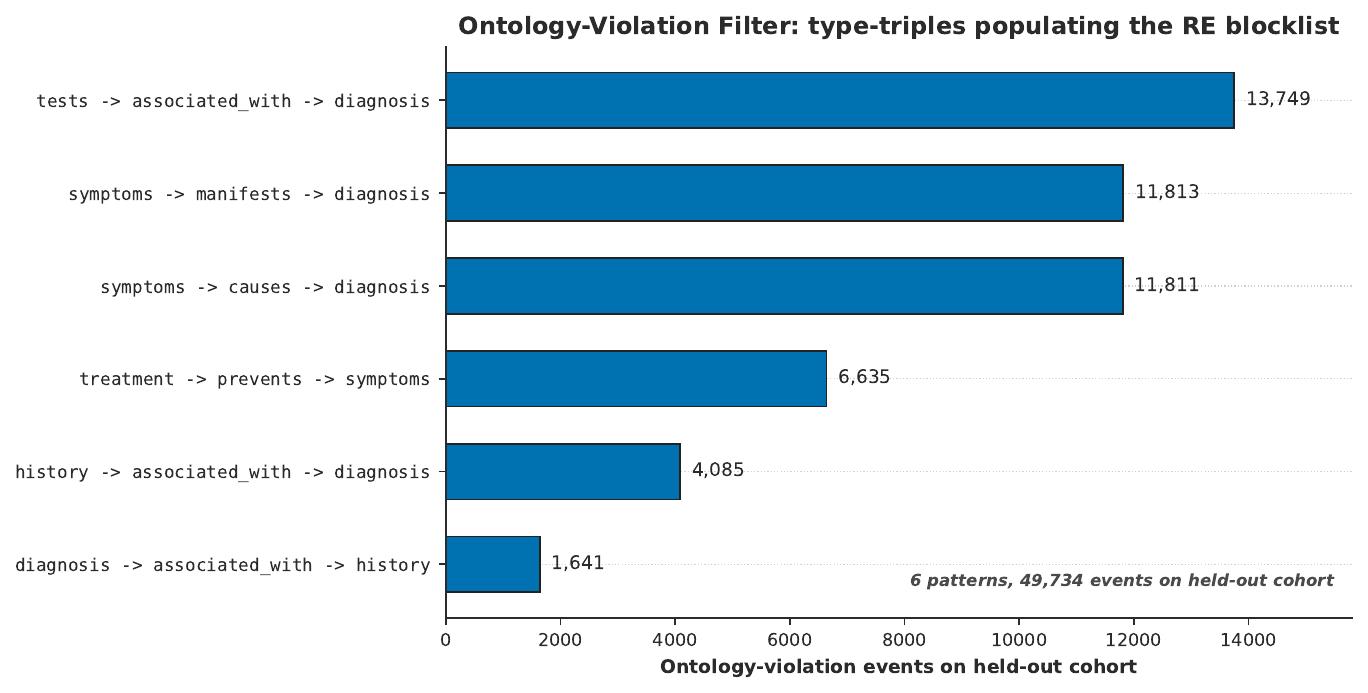}
    \caption{Type-triple patterns populating the relation blocklist through the
    Ontology-Violation Filter (Section~\ref{sec:design-ontology}) on the held-out
    cohort. Six threshold-clearing patterns aggregate 49{,}734 ontology-violation
    events. The Filter is a deterministic predicate
    (Equation~\ref{eq:ontology}) over head-type, relation, tail-type triples against
    a 29-entry clinical ontology, independent of the verifier-fed pathway whose
    collapse is documented in Figure~\ref{fig:f1-channel}.}
    \label{fig:ontology}
\end{figure}

\paragraph{Suppression magnitude.}
The Filter silently suppressed $N = 1{,}207$ relation candidates (0.98\% of total):
the ON arm carried 122{,}253 relations against 123{,}460 in the OFF arm, all
1{,}207 removed through the verifier-independent pathway. Where the verifier-fed
pathway produced zero usable patterns at production scale
(Section~\ref{sec:results-f1}), the Filter populates the same downstream blocklist
through an independent signal source that sidesteps the persistence-decay
interaction. Figure~\ref{fig:ontology} shows the six threshold-clearing patterns.

\begin{sloppypar}
\paragraph{Mechanism stability.}
The same six type-triple patterns recur on the held-out cohort with the same
rank-ordering and proportional frequencies as on the development cohort
(\texttt{tests-associated\_with-diagnosis}, \texttt{symptoms-manifests-diagnosis},
\texttt{symptoms-causes-diagnosis}, \texttt{treatment-prevents-symptoms},
\texttt{history-associated\_with-diagnosis},
\texttt{diagnosis-associated\_with-history}). The total event count grows from
2{,}486 on the development cohort to 49{,}734 on the held-out cohort, a 20.0$\times$
increase over a 25$\times$ increase in patients. This sub-linear growth is the
signature of a signal concentrating on a fixed type-triple alphabet rather than
dispersing. Each pattern is, by construction, a type-triple the ontology disallows;
whether individual suppressions improve dataset quality is a downstream question
outside this paper's scope.
\end{sloppypar}

\paragraph{No precision-proxy or runtime delta.}
The verifier passed essentially every relation reaching enriched output in both
arms (precision proxy $1.0000$ in each), and the wall-time difference was 26 seconds
across runs of roughly nine hours (0.08\%, within noise). The Filter's contribution
is therefore upstream candidate suppression through an independent signal source,
not a downstream precision gain, in a regime where the verifier itself is permissive.

\paragraph{A note on provenance.}
The 1{,}207 suppressions here are the Filter's isolated contribution, measured
against an OFF arm that kept the NER-coverage gate ON. The combined ablation of
Section~\ref{sec:results-downgrade} reports a smaller figure of 913 because its OFF
arm disables every DBPM mechanism, including the NER-coverage gate. Both are
correct; they answer different questions, and we return to the distinction in
Section~\ref{sec:results-downgrade}.

\subsection{Five Signal Sources for the Question-Answering Gate}
\label{sec:results-iterations}

\begin{figure}[t]
    \centering
    \includegraphics[width=\linewidth]{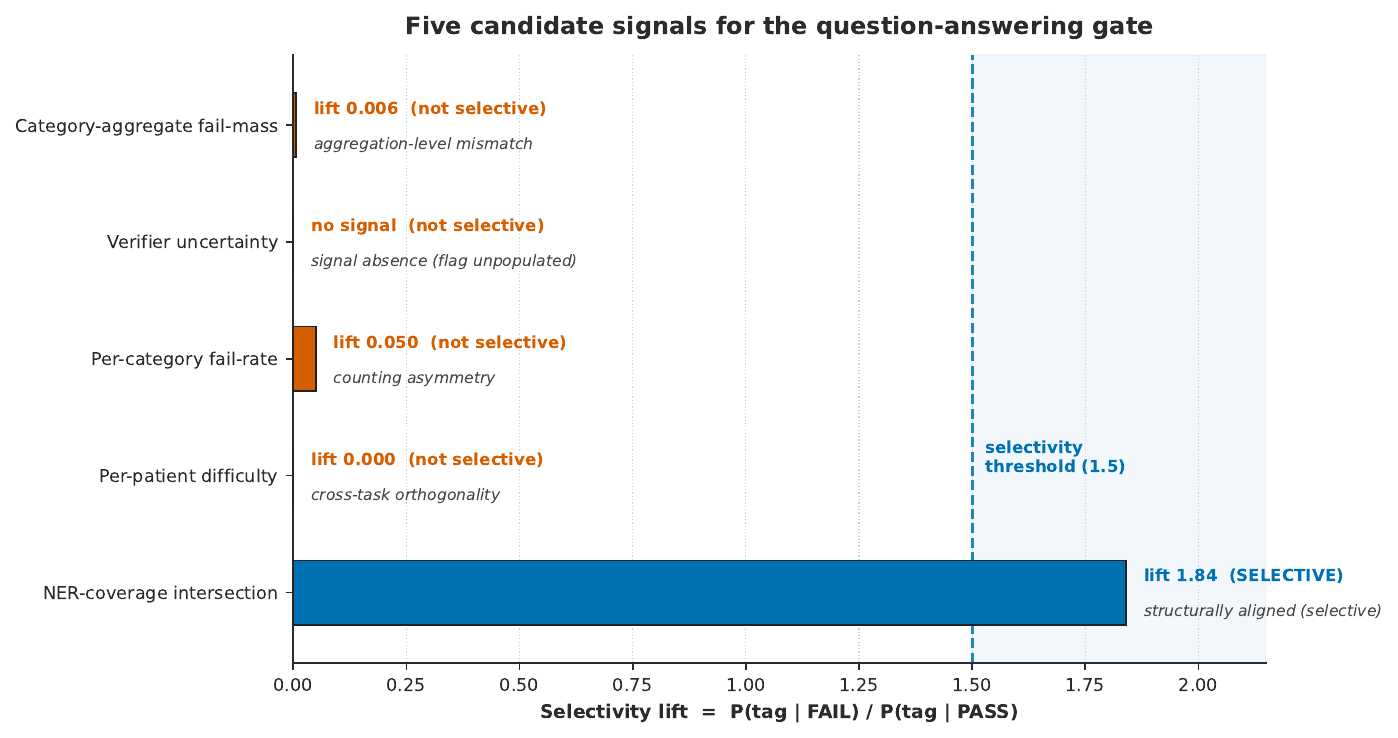}
    \caption{Selectivity lift of the five candidate signals for the question-answering gate on
the held-out 5{,}000-patient cohort. Lift is defined in Equation~(\ref{eq:lift}) as
$P(\mathrm{tag} \mid \mathrm{FAIL}) / P(\mathrm{tag} \mid \mathrm{PASS})$; values
$\geq 1.5$ (dashed line) indicate selectivity. Four of the five signals fall below
the threshold, each for the distinct reason labelled; the NER-coverage signal reaches
a path lift of 1.84 in the full-DBPM configuration, within the 1.52--1.84 empirical
band across four independent replications (Section~\ref{sec:limits-construct}).
}
\label{fig:iterations}
\end{figure}

We tested five candidate signals for the question-answering gate, that is, five
different quantities the gate could read to make its decision, in the order
introduced in Section~\ref{sec:design-qar}. Four are not selective, each failing for
a structurally distinct reason; the fifth, the NER-coverage gate, is selective
within its operating envelope. Figure~\ref{fig:iterations} plots the selectivity
lift of all five.

\paragraph{Category-aggregate fail-mass.}
The first source totals each category's accumulated rejection mass and flags
categories above a threshold. On the development cohort, its lift was 0.006 across
$N = 2000$ pairs: not merely non-selective but anti-selective, tagging categories
roughly 170$\times$ \emph{less} likely to be verifier-rejected than untagged ones.
The cause is an \emph{aggregation-level mismatch}: the signal decides at the category
level whereas the verifier decides per pair, so it flags whichever categories
accumulate volume rather than those that are error-prone.

\paragraph{Verifier uncertainty.}
The second source reads the per-pair field the verifier sets when its output fails
to parse, hypothesised to mark uncertainty. Across 1{,}283 development-cohort pairs,
no event carried an ambiguous value: the signal does not exist at this scale. This
is a case of \emph{signal absence}: the field is a parse-failure flag, and modern
instruction-tuned models almost never produce unparseable output, so it is left
unset.

\paragraph{Per-category fail-rate.}
The third source estimates each category's true fail-rate
(Equation~\ref{eq:m1v5-failrate}) by joining rejection and success records, firing a
downgrade above a fail-rate threshold. The fail-rate genuinely varies across
categories (spread 0.52 on the held-out cohort, 0.58 on the development cohort, so
the signal is non-degenerate), yet the gate's instance-level lift was only 0.05. The
cause is a \emph{counting asymmetry}: failures are counted per pair while successes
are counted per question template (about eleven templates total), so the estimate
overstates the true rate for high-volume categories and the gate tags categories
whose individual pairs mostly pass.

\paragraph{Per-patient difficulty.}
The fourth signal, the first to read cross-task state at decision time, combines the
category fail-rate with a per-patient upstream-difficulty score
(Equation~\ref{eq:m1v6-difficulty}) and fires only when both exceed threshold. Its
lift was 0.00, a failure we attribute to \emph{cross-task orthogonality}: the
PASS-side and FAIL-side difficulty distributions are statistically indistinguishable
(medians 0.179 versus 0.174, means 0.188 versus 0.191, third quartiles 0.290 versus
0.294). Upstream extraction difficulty simply does not predict verifier rejection at
the answer stage, which turns on answer content such as hallucination or
over-specification.

\paragraph{NER-coverage intersection.}
The fifth source asks whether the patient has the extracted entities an answer in
the asked category would need (Section~\ref{sec:design-qar}). It was motivated by a
read-only test on the development cohort: across six qualified categories, a patient
lacking the relevant coverage was a median 3.43$\times$ more likely to have its
answer rejected than a patient with it. The deployed gate allows covered patients
and downgrades uncovered ones, across the nine categories with a matching NER bucket;
the two without a bucket (\texttt{adverse\_event}, \texttt{outcome\_mortality}) fall
through to the difficulty gate.

On the development cohort, the gate's path lift across the nine mapped categories was
1.89 (tagging 0.386 of FAIL pairs against 0.205 of PASS pairs); the two unmapped
categories contributed a fallback lift of 0.00, giving a global lift of 1.59. On the
held-out cohort, the path lift was 1.84 (0.360 of FAIL against 0.196 of PASS, over
$n_{\mathrm{FAIL}} = 3{,}868$ and $n_{\mathrm{PASS}} = 31{,}135$ mapped-category
pairs),\footnote{The lift uses $n_{\mathrm{PASS}} = 31{,}135$, every PASS pair
evaluated by the gate. A second count, 17{,}703 covered-PASS pairs, appears later in
this section and restricts to PASS pairs whose patient-level NER set covers the
category; the two are not interchangeable.} with the same 0.00 fallback giving a
global lift of 1.52. The path lift therefore holds across the 25$\times$ scale-up to
within roughly 4\% (1.89 to 1.84), although the underlying tagging probabilities move
somewhat more, which we note rather than overclaim.

The held-out path lift is statistically significant. We obtain its 95\% confidence
interval from the delta-method variance for the log of a risk ratio,
\begin{equation*}
\widehat{\mathrm{Var}}[\log \mathrm{lift}] =
   \frac{1 - P_{\mathrm{tag}|\mathrm{FAIL}}}{n_{\mathrm{FAIL}} \cdot P_{\mathrm{tag}|\mathrm{FAIL}}}
 + \frac{1 - P_{\mathrm{tag}|\mathrm{PASS}}}{n_{\mathrm{PASS}} \cdot P_{\mathrm{tag}|\mathrm{PASS}}},
\label{eq:lift-variance}
\end{equation*}
which gives $[1.75, 1.93]$; the null hypothesis of no selectivity
($\mathrm{lift} = 1$) is rejected decisively ($p < 10^{-15}$). This interval captures
within-run sampling precision only; run-to-run and configuration-dependent variation
is reported separately in Section~\ref{sec:limits-construct}.

\emph{Mechanism integrity.} Of the 17{,}703 covered-PASS pairs in mapped categories
on the held-out cohort, zero were tagged, a structural-correctness check confirming
that covered pairs hit the early-ALLOW exit and never reach the downgrade branch.
The development-cohort check (726 covered-PASS pairs, zero tagged) replicates this at
25$\times$ smaller scale.

\paragraph{Why the fifth source succeeds where the first four fail.}
The four failures are not a ladder of increasing sophistication that eventually
works; the second and fourth sources are more elaborate than the first and fail more
conclusively. They fail because each reads the wrong input. The first and third base their
decision on category-level statistics, but the verifier decides per pair. The second reads a
signal that does not exist at scale. The fourth reads upstream difficulty, which is
orthogonal to what the verifier judges. The fifth reads whether the patient has the
entities an answer would need to be grounded, which is the same question the verifier
implicitly resolves. The structural pattern is the subject of
Section~\ref{sec:discussion-principle}: a pre-generation signal is selective for
verifier rejection only when it directly probes the answer-grounding question the
verifier itself evaluates.

\subsection{Per-Category Operating Envelope}
\label{sec:results-envelope}

\begin{figure}[t]
    \centering
    \includegraphics[width=\linewidth]{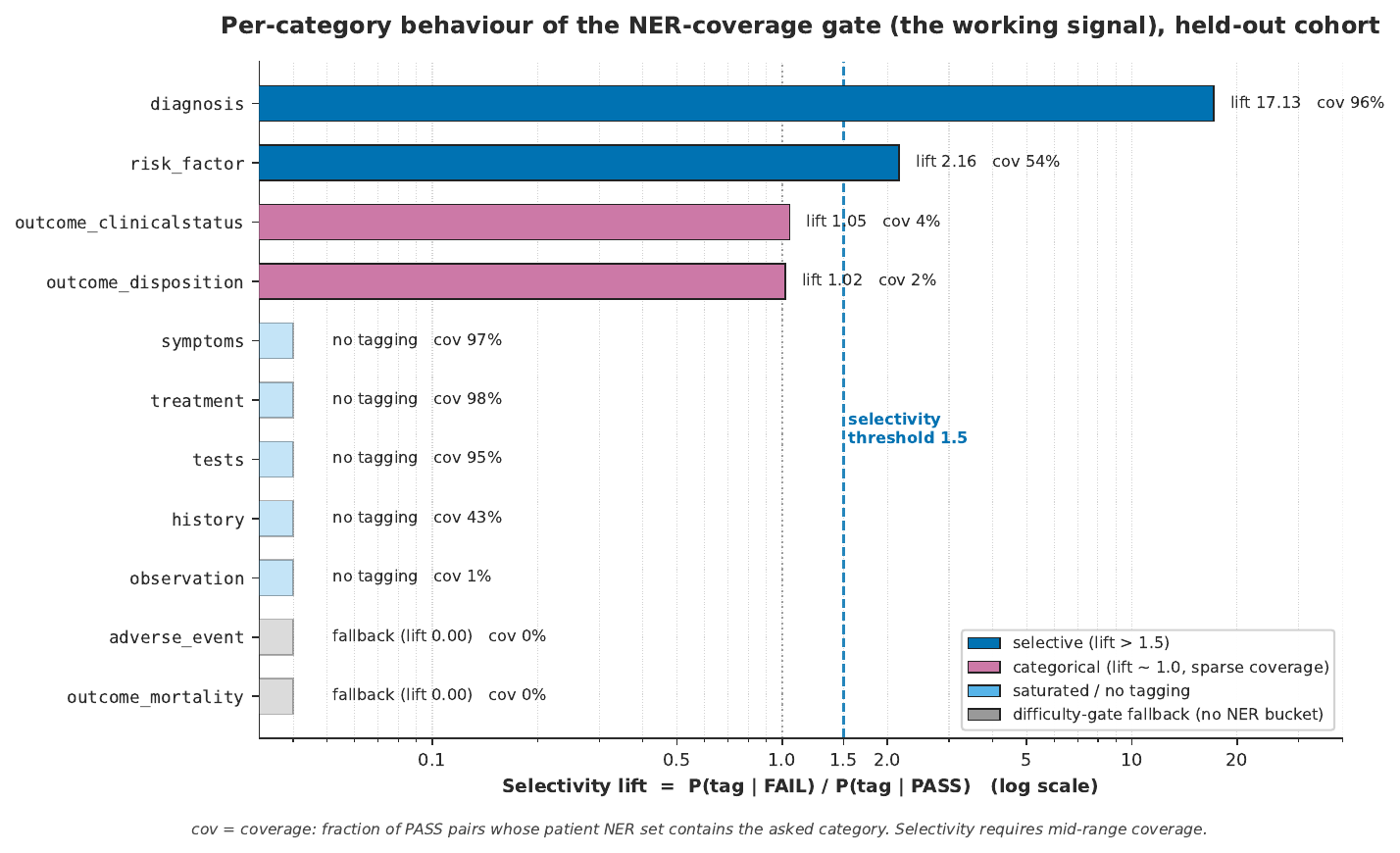}
    \caption{Per-category behaviour of the NER-coverage gate on the held-out cohort,
the working signal of Section~\ref{sec:results-iterations}. Bars give the
selectivity lift per QAR category on a log scale; coverage (the fraction of PASS
pairs whose patient-level NER set contains the asked category) is annotated on each.
The categories fall into four regimes: selective (\texttt{diagnosis} 17.13,
\texttt{risk\_factor} 2.16, both above the 1.5 threshold); categorical at sparse
coverage (\texttt{outcome\_clinicalstatus} 1.05, \texttt{outcome\_disposition}
1.02); saturated or no-tagging at high or very low coverage (five categories); and
difficulty-gate fallback where no NER bucket exists (\texttt{adverse\_event},
\texttt{outcome\_mortality}). Selectivity requires mid-range coverage:
\texttt{diagnosis} is anti-selective in isolation but becomes the most selective
category under full DBPM, as the Ontology-Violation Filter removes ontology-violating
relations that inflate its coverage signal (Section~\ref{sec:results-iterations}).}
\label{fig:envelope}
\end{figure}

The headline path lift of 1.84 averages over substantial per-category variation;
selectivity is far from uniform across the eleven categories.
Table~\ref{tab:per-category} gives the full breakdown and Figure~\ref{fig:envelope}
visualises it. Four operating regimes emerge.

\begin{table}[h]
\centering
\caption{Per-category selectivity of the NER-coverage gate on the held-out cohort
(single-run measurement; per-category lifts for categories with small
$n_{\mathrm{FAIL}}$ are subject to run-to-run variance, discussed at the end of this
subsection). The coverage column is the percentage of PASS pairs whose patient-level
NER set covers the category. Categories with no NER bucket fall through to the
difficulty gate. ``Regime'' is the operating-envelope classification described in the
body.}
\label{tab:per-category}
\small
\renewcommand{\arraystretch}{1.15}
\begin{tabular}{lrrrrrrl}
\hline
\textbf{Category} & $n_{\mathrm{PASS}}$ & $n_{\mathrm{FAIL}}$ & \textbf{cov\%} & $P(t \mid P)$ & $P(t \mid F)$ & \textbf{lift} & \textbf{regime} \\
\hline
\texttt{diagnosis}               & 4{,}764 & 105 & 95.7 & 0.043 & 0.733 & \textbf{17.13} & selective \\
\texttt{risk\_factor}            &   564 & 448 & 53.7 & 0.463 & 1.000 & \textbf{2.16} & selective \\
\texttt{outcome\_clinicalstatus} & 3{,}191 & 448 &  4.4 & 0.956 & 1.000 &       1.05  & categorical \\
\texttt{outcome\_disposition}    & 2{,}235 & 419 &  1.7 & 0.983 & 1.000 &       1.02  & categorical \\
\texttt{outcome\_mortality}      &   697 & 351 &  0.0 (no-NER) & 0.149 & 0.000 & 0.00 & fallback \\
\texttt{adverse\_event}          & 2{,}517 & 443 &  0.0 (no-NER) & 0.113 & 0.000 & 0.00 & fallback \\
\texttt{history}                 & 1{,}971 & 433 & 42.8 & 0.000 & 0.000 & undefined & no tagging \\
\texttt{observation}             & 2{,}979 & 416 &  1.3 & 0.000 & 0.000 & undefined & no tagging \\
\texttt{symptoms}                & 3{,}944 & 303 & 96.8 & 0.000 & 0.000 & undefined & saturated \\
\texttt{treatment}               & 4{,}143 & 260 & 97.7 & 0.000 & 0.000 & undefined & saturated \\
\texttt{tests}                   & 4{,}130 & 242 & 94.8 & 0.000 & 0.000 & undefined & saturated \\
\hline
\end{tabular}
\end{table}

\paragraph{Selective (lift $> 1.5$).}
Two categories separate PASS from FAIL. \texttt{diagnosis} (lift 17.13) is the
strongest: at 95.7\% coverage the gate tags 73.3\% of FAIL pairs against only 4.3\%
of PASS pairs, a clean separation driven by a small, well-separated FAIL stratum (105
FAIL against 4{,}764 PASS). \texttt{risk\_factor} (lift 2.16) sits in the
mid-coverage regime the gate was designed for: at 53.7\% coverage it tags every FAIL
pair and 46\% of PASS pairs. Notably, \texttt{diagnosis} is anti-selective when the
fifth source is ablated in isolation but becomes the most selective category under
full DBPM, because the Ontology-Violation Filter removes ontology-violating relations
that would otherwise inflate the coverage signal for diagnosis questions; cleaning
the upstream relations sharpens the signal. We examine this interaction in
Section~\ref{sec:discussion-principle}.

\paragraph{Categorical (lift $\approx 1.0$).}
Two outcome categories, \texttt{outcome\_clinicalstatus} (coverage 4.4\%, lift 1.05)
and \texttt{outcome\_disposition} (coverage 1.7\%, lift 1.02), tag nearly all PASS
and all FAIL pairs at near-zero coverage, behaving as coarse always-on category
filters rather than per-patient gates. They are not selective, but their behaviour is
predictable.

\paragraph{No tagging or saturated.}
Five of the general clinical categories, \texttt{history}, \texttt{observation},
\texttt{symptoms}, \texttt{treatment}, and \texttt{tests}, tag nothing. For \texttt{symptoms},
\texttt{treatment}, and \texttt{tests}, coverage is saturated (95--98\%), so nearly
every patient is covered and the gate allows almost everyone. For \texttt{history}
(coverage 42.8\%) and \texttt{observation} (coverage 1.3\%), the gate simply does not
fire on this cohort. The mechanism operates correctly but rarely activates.

\paragraph{Fallback (no NER bucket).}
\texttt{adverse\_event} and \texttt{outcome\_mortality} have no matching NER bucket,
so the fifth source cannot apply and the difficulty gate runs, tagging PASS pairs
only ($P(t \mid F) = 0$).

The net picture is an operating envelope rather than a uniform effect: strong
per-patient selectivity in two categories, coarse categorical filtering in two, no
measurable tagging in five, and no application in two. The gate is selective in the
mid-to-high-coverage regime where the coverage signal separates the PASS and FAIL
strata, and degenerates predictably outside it. Identifying these conditions is
itself a methodological contribution.

\paragraph{Per-category instability for small FAIL strata.}
The per-category lifts for categories with small FAIL strata (\texttt{diagnosis},
$n_{\mathrm{FAIL}} = 105$; the \texttt{outcome} categories, $n_{\mathrm{FAIL}} \leq
451$; \texttt{adverse\_event}, $n_{\mathrm{FAIL}} = 447$) are subject to run-to-run
noise. An independent replication under the full-DBPM configuration (the cross-task
ablation arm, Appendix~\ref{app:crosstask}) gave a path lift of 1.71, within the
1.52--1.84 band, but the per-category \texttt{diagnosis} lift swung between near-zero
and large positive values across runs and configurations of the same gate. We caution
against drawing per-category conclusions from any single run: the aggregate path lift
is stable, the per-category lifts on small strata are not. The regime-level
classification (selective, categorical, saturated, fallback) is robust across runs
even where the precise within-regime lifts vary.

\subsection{Tagging, Not Suppression: Measurement Discipline and Provenance}
\label{sec:results-downgrade}

\begin{figure}[t]
    \centering
    \includegraphics[width=\linewidth]{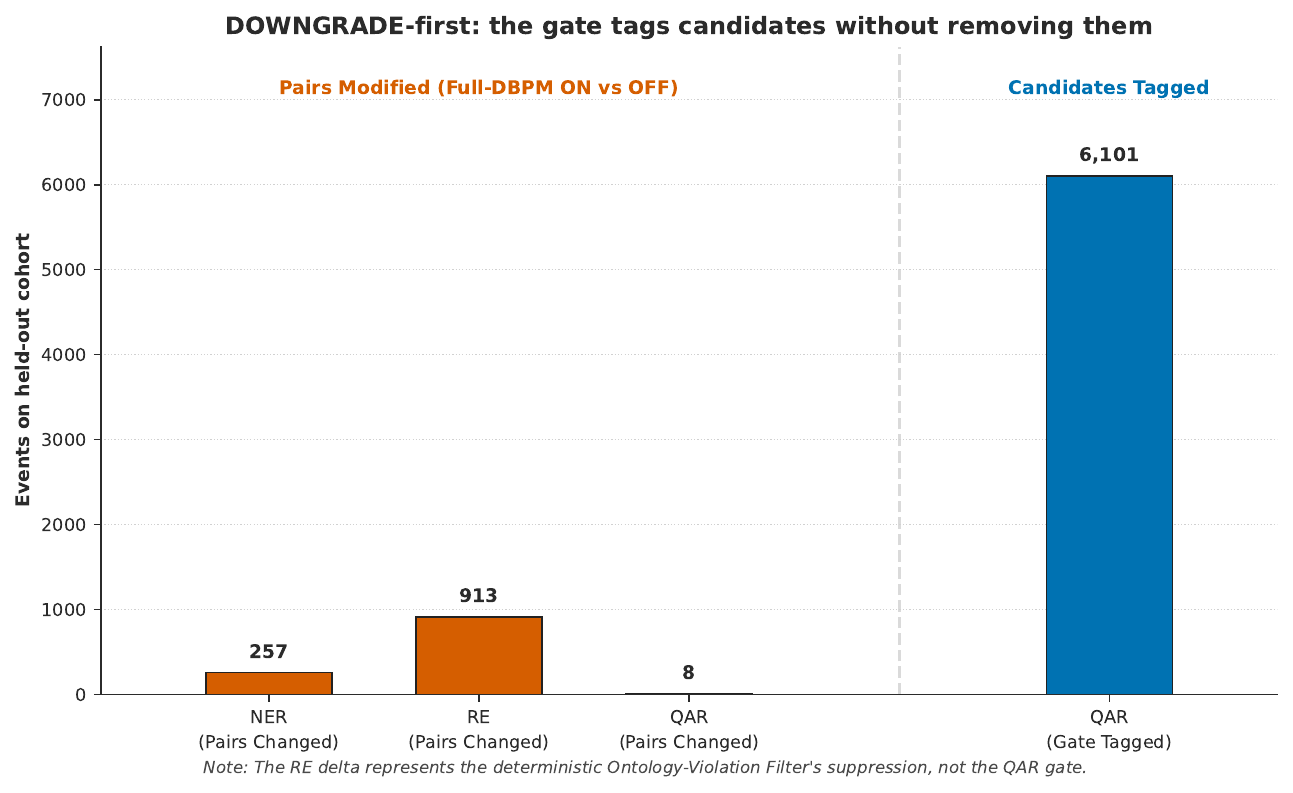}
    \caption{Pair-count deltas (left) versus gate-tag counts (right) on the held-out
    cohort under the full-DBPM ablation. The named-entity and QA pair-count deltas
    are within run-to-run noise ($+257$ and $-8$), confirming the DOWNGRADE-first
    discipline: the gate tags QA candidates without removing them. The relation delta
    of $-913$ is the Ontology-Violation Filter's genuine suppression, not a
    gate effect. The gate tags 6{,}101 PASS-side and 1{,}392 FAIL-side QA pairs
    (7{,}493 total), establishing selectivity at the tagging level rather than through
    output-volume change.}
    \label{fig:downgrade-first}
\end{figure}

Under the DOWNGRADE-first discipline, the NER-coverage gate tags pairs without
removing them, so the gate produces almost no change in pair counts even while
tagging thousands of pairs. Table~\ref{tab:v7-pair-deltas} shows the pair-count
deltas and Figure~\ref{fig:downgrade-first} contrasts them against the tagging
volume.

\begin{table}[h]
\centering
\caption{Pair-count deltas between the full-DBPM ON and OFF arms on the held-out
cohort. The named-entity ($+0.19\%$) and QA ($-0.03\%$) deltas are within the
per-stage non-determinism floor, confirming that the NER-coverage gate tags QA
candidates without removing them. The relation delta ($-913$, $-0.74\%$) is the
Ontology-Violation Filter's genuine suppression, the only above-noise pair-count
change.}
\label{tab:v7-pair-deltas}
\small
\renewcommand{\arraystretch}{1.15}
\begin{tabular}{lrrrr}
\hline
\textbf{Stage} & \textbf{ON} & \textbf{OFF} & $\boldsymbol{\Delta}$ & $\boldsymbol{\Delta\%}$ \\
\hline
NER & 133{,}790 & 133{,}533 & $+257$ & $+0.19\%$ \\
RE  & 122{,}425 & 123{,}338 & $-913$ & $-0.74\%$ \\
QAR &  31{,}135 &  31{,}143 &   $-8$ & $-0.03\%$ \\
\hline
\end{tabular}
\end{table}

The gate tagged 6{,}101 verifier-PASS QA pairs in the ON arm against zero in the OFF
arm (where the master switch deactivates every mechanism), so all 6{,}101 PASS-side
tags are attributable to DBPM. With the 1{,}392 FAIL-side tags, 7{,}493 QA pairs are
tagged in total, the basis of the confusion summary in Table~\ref{tab:v7-confusion}.
Against this tagging volume, the near-zero pair-count deltas confirm the design
intent: selectivity must be read from tagging behaviour, not output volume.

\begin{table}[H]
\centering
\caption{Tagging summary on the held-out cohort (DBPM-full ablation) across all
eleven QA categories. Rows are the verifier's verdict; columns are the gate's tag
(DOWNGRADE collapsed with BLOCK as \emph{tagged}). Percentages are within-row, the
fraction of each verdict class that the gate tags, so each row sums to 100\%. The
table reports selectivity, not classification accuracy: the gate is a selective
tagger on a single grounding signal, not a verifier surrogate, so the relevant
quantity is the ratio of the two tag rates (the lift), not precision, recall, or
F1. The gate tags 36.0\% of verifier-FAIL pairs against 19.6\% of verifier-PASS
pairs, a lift of $0.360 / 0.196 = 1.84$ (95\% CI $[1.75, 1.93]$). Tagging in the
nine mapped categories is produced by the NER-coverage gate; the two no-bucket
categories (\texttt{adverse\_event}, \texttt{outcome\_mortality}) are tagged by the
difficulty gate, which contributes no FAIL-side tags, so the all-category ratio
equals the mapped-category path lift, an equality specific to these data, not a
general identity.}

\label{tab:v7-confusion}
\small
\renewcommand{\arraystretch}{1.2}
\begin{tabular}{lrrr}
\hline
\textbf{Verifier verdict} & \textbf{Tagged} & \textbf{Not tagged} & \textbf{Row total} \\
\hline
FAIL & 1{,}392 (36.0\%) & 2{,}476 (64.0\%) & 3{,}868 \\
PASS & \phantom{0}6{,}101 (19.6\%) & 25{,}034 (80.4\%) & 31{,}135 \\
\hline
\textbf{Column total} & \textbf{7{,}493} & \textbf{27{,}510} & \textbf{35{,}003} \\
\hline
\end{tabular}
\end{table}

\paragraph{Aggregate DBPM provenance (single combined ablation).}
A single paired ablation toggles all new DBPM mechanisms together via the master
switch. On the same held-out cohort it produced 913 silent relation suppressions by
the Ontology-Violation Filter (ON arm 122{,}425 relations against 123{,}338 in the
OFF arm) and 6{,}101 attributable QA downgrade-tags by the NER-coverage gate (against
zero when the switch is off). Together these are 7{,}014 gating events across the
287{,}350 pipeline pairs of the ON arm, or 2.44\% of pipeline pairs touched by the
new DBPM mechanisms.

The combined-ablation suppression count of 913 differs from the isolated count of
1{,}207 (Section~\ref{sec:results-ontology}) because the two ablations differ in
\emph{both} arms, not only the OFF arm: the isolated ablation holds the NER-coverage
gate ON in both arms and measures only the Filter, whereas the combined ablation
turns every mechanism off together, which shifts the relation pair counts in both
arms as well. The two numbers are not in conflict; each is the right answer to a different
question. The combined figure of 913 answers ``What is DBPM's aggregate contribution
against an ungated baseline?''; the isolated figure of 1{,}207 answers ``What is the
Filter's contribution holding everything else constant?''

At this scale, the allow-set dominates the blocklist: the named-entity channel holds
96{,}708 whitelisted signatures against 2{,}991 blocklist patterns, and the QA
channel holds 11 whitelisted signatures against 34 blocklist patterns. This dominance is consistent with
the early-ALLOW exit of Equation~(\ref{eq:gate}): most candidates match a known-safe
signature and bypass belief-based gating.

\paragraph{Runtime overhead.}
The ON arm ran in 8h48m10s (6.34\,s/patient) and the OFF arm in 8h48m50s
(6.35\,s/patient), a difference of 40 seconds across 31{,}690, or 0.13\%. This is
below the pipeline's per-patient variance floor (expected noise on the mean across
$n = 5000$ is roughly $\pm 140$ seconds; continuous batching, the dual-model
verifier, and network jitter contribute 1--3 seconds of per-patient standard
deviation). We therefore claim only that the gate's per-patient overhead is below the
variance floor, not that it produces a speed-up.

Two consequences follow for downstream consumers. First, DBPM's gate is
measurement-instrumented rather than production-suppressive: it records which pairs
\emph{would} be blocked under hard-BLOCK mode without committing to the removal, and a
deployment wishing to act on its verdicts would enable hard-BLOCK mode
(Section~\ref{sec:design-downgrade}). Second, in Table~\ref{tab:v7-pair-deltas} the
named-entity and QA deltas are within non-determinism, and the relation delta of
$-913$ is the Filter's suppression, not a gate effect; the gate itself, in
DOWNGRADE-first mode, removes no pairs. Conclusions from DBPM ablations must therefore
be drawn at the tagging level, not from pair-count deltas.

\paragraph{Provenance scope.}
The 2.44\% figure comes from a single master-switch ablation on the locked cohort,
measuring the aggregate contribution of the two new mechanisms against an ungated
baseline on identical input. Pattern-level cross-task propagation is disabled in both
arms by the master switch; its independent contribution is measured separately in
Appendix~\ref{app:crosstask} at 0.098\% of pipeline throughput, below the per-stage
non-determinism floor.

\section{Discussion}
\label{sec:discussion}

This section sets out what the results of Section~\ref{sec:results} establish about
inference-time pattern-memory gating in LLM-grounded clinical NLP. We organise it
around three questions: what the five-source dissection of the question-answering
gate reveals as a structural principle (Section~\ref{sec:discussion-principle}); how
the production-scale failure of the verifier-fed channel changes the design problem
for downstream pipelines (Section~\ref{sec:discussion-failure-mode}); and what
follows for the safe deployment of LLM-driven extraction at scale
(Section~\ref{sec:discussion-clinical}).

\subsection{A Structural Principle for Selective Pre-Generation Gating}
\label{sec:discussion-principle}

The five-signal dissection of the question-answering gate
(Sections~\ref{sec:design-qar} and~\ref{sec:results-iterations}) is enough to state
a structural principle about which signals can make inference-time gating selective
in a generator-verifier architecture. In each of the first four gate variants, the
gate read a structurally distinct signal and failed for a structurally distinct
reason; the fifth succeeded by reading a signal none of the others did.
Table~\ref{tab:iteration-signals} lays out the progression.

\begin{table}[h]
\centering
\caption{The five signal sources tested for the question-answering gate,
characterised by what each read, its granularity, the empirical outcome, and the
failure mode where applicable. The NER-coverage gate is the working design; its path
lift band of 1.52--1.84 is measured across four independent replications
(Section~\ref{sec:limits-construct}).}
\label{tab:iteration-signals}
\small
\renewcommand{\arraystretch}{1.25}
\begin{tabular}{p{2.4cm}p{3cm}p{2cm}p{2.4cm}p{3.6cm}}
\hline
\textbf{Source} & \textbf{Signal read} & \textbf{Granularity} & \textbf{Outcome} & \textbf{Failure mode} \\
\hline
Category-aggregate fail-mass & category-aggregate verifier fail-mass & category-level & anti-selective (lift 0.006) & aggregation-level mismatch \\
Verifier uncertainty & verifier parse-failure flag & per-pair & signal absent & signal absence \\
Per-category fail-rate & per-category fail-rate (dual-bucket) & category-level & anti-selective (lift 0.05) & counting asymmetry \\
Per-patient difficulty & per-patient upstream difficulty & per-patient, cross-task & no signal (lift 0.00) & cross-task orthogonality \\
NER-coverage intersection & per-patient NER-category coverage & per-patient, cross-task & selective (path lift 1.52--1.84 across four replications) & working design \\
\hline
\end{tabular}
\end{table}

Read across the five, the pattern is \emph{not} that more sophisticated mechanisms
win. The second and fourth sources are more elaborate than the first, and they fail
more conclusively. The pattern is that a pre-generation signal is selective for
verifier rejection only when it directly probes the answer-grounding question the
verifier itself evaluates. The first and third gate variants test a category-level statistic, the accumulated
fail-mass or fail-rate of the whole category, but the verifier accepts or rejects
each pair individually, so a category-level signal cannot separate the pairs the
verifier distinguishes within that category. The second variant checks a
parse-failure flag, on the assumption that unparseable output marks an uncertain
verdict, but modern instruction-tuned models almost always produce parseable output,
so the flag is essentially never set and carries no information. The fourth variant
checks how much upstream extraction the patient required, on the assumption that
hard-to-extract patients yield worse answers; but the verifier judges the answer's
own content, whether it is hallucinated, over-specified, or unsupported by the source
text, and this is unrelated to upstream extraction difficulty. We confirmed the two
are unrelated directly: the upstream-difficulty scores of verifier-passed and
verifier-failed pairs are statistically indistinguishable
(Section~\ref{sec:results-iterations}). The fifth variant checks whether the patient
has the extracted clinical entities that an answer in the asked category would need,
which is the same question the verifier resolves when it decides whether an answer is
grounded. That alignment between the gate's signal and the verifier's decision
criterion is what makes the fifth variant selective.

The principle is directly usable. A practitioner designing an inference-time gate
for a generator-verifier pipeline can test a candidate signal by asking whether it
probes the same question the verifier answers. If it does, the gate may be selective;
if it does not, the gate will not be, however sophisticated its aggregation. The
five sources give one positive and four negative instances of exactly this test.

Alignment is necessary for selectivity but does not guarantee it. The four failed
variants establish the necessary half: none was selective, and none probed the
verifier's grounding question. But alignment alone is not sufficient, a signal can
probe the right question and still fail to separate PASS from FAIL when its values
do not vary across patients. The per-category operating envelope
(Section~\ref{sec:results-envelope}) shows this: even the well-aligned NER-coverage
signal degenerates under coverage saturation, where nearly every patient has the
relevant entities and the gate rarely fires, and under coverage sparsity, where
nearly no patient has them and the gate fires categorically rather than per patient.
Signal-distributional properties bound the operating regime beyond the alignment requirement.

The principle also explains an interaction between the Ontology-Violation Filter and
the NER-coverage gate. In isolation, the \texttt{diagnosis} category is
anti-selective, but under full DBPM it becomes the most selective category (lift
17.13, Section~\ref{sec:results-envelope}). The Filter removes ontology-violating
relations that would otherwise enter the coverage signal the gate reads; with those
spurious relations gone, the coverage signal aligns more closely with the verifier's
grounding decision for diagnosis questions. This is the principle operating one level
up: cleaning an upstream signal makes a downstream signal more aligned with the
verifier's question, and so more selective.

\subsection{Verifier-Fed Memory as a Production-Scale Failure Mode}
\label{sec:discussion-failure-mode}

The empty verifier-fed relation-extraction channel at production scale
(Section~\ref{sec:results-f1}), zero persisted patterns against 785{,}797 logged
classifier rejections, is a production-scale failure of the most natural design for
an inference-time pattern memory. That design accumulates per-signature rejection
statistics through a source-weighted update (Equation~\ref{eq:severity-update}),
applies time-decay so the memory can recover as documentation conventions shift
(Equation~\ref{eq:wallclock-decay}), and prunes signatures whose decayed severity
falls below a floor. Each mechanism is individually sensible; their interaction at
production scale is not.

What determines the outcome is the distributional shape of the rejection stream, not
its volume. When rejections concentrate, many failures sharing a few signatures, each
signature is reinforced faster than it decays and clears threshold. When rejections
disperse, many failures spread thinly across many signatures, each signature decays
between save events faster than reinforcement arrives, and nothing clears threshold
no matter how high the total volume. Across the full 167{,}034-patient run, the relation rejection
stream was dispersed: its 785{,}797 rejections spread across the open
natural-language vocabulary of relation labels the model emits, with no single
signature reinforced fast enough to survive.

The Ontology-Violation Filter (Section~\ref{sec:design-ontology}) avoids this failure
by changing the source of the signal that feeds the channel, not the persistence mechanism. Its predicate
fires per candidate against a fixed set of 29 type-triples, so events concentrate on
those triples regardless of how the relation label is phrased. The concentrated
distribution clears threshold where the dispersed one did not, and it does so through
the same persistence machinery; only the input signal differs. The contribution here
is the empirical demonstration that the choice of input signal, the verifier's verdicts versus the ontology predicate, not the design of
the memory, can be the difference between an empty channel and a populated one across the full 167{,}034-patient run.

The implication reaches past this pipeline. Any inference-time memory that
accumulates per-signature statistics over free-form LLM output, whether relation
labels, classification tags, or intent labels, is exposed to the same dispersion. The
remedy is to define signal sources that map that free-form output onto a bounded
canonical space, so the channel concentrates rather than disperses.

\subsection{Clinical Implications}
\label{sec:discussion-clinical}

Three findings of this paper carry directly into clinical deployment practice: a
pattern-memory blocklist can be configured yet empirically empty at scale, an
ontology signal and a verifier catch different error classes and so complement rather
than replace each other, and tagging suspect candidates rather than deleting them
keeps every gating decision in the clinical audit trail. We state each as an
implication for practitioners adopting inference-time pattern-memory gating in
LLM-grounded extraction pipelines.

\paragraph{(i) A blocklist that appears configured may be empirically inactive.}
Section~\ref{sec:results-f1} documents a verifier-fed relation blocklist that was
empty at production scale even though the pipeline generated nearly 800{,}000
rejections that should, in principle, have filled it. A pipeline that relied on this
blocklist for a safety guarantee would have shipped without that guarantee being
enforceable on its full corpus. The lesson is that a deployment using a pattern-memory
gate needs observability for whether the memory \emph{actually populates} at scale,
not merely confirmation that the mechanism is configured to populate it. We release
the \texttt{crosscheck\_F1\_167k\_blocklist.py} script as one such check.

\paragraph{(ii) Verifier-independent signal sources complement rather than replace
the verifier.}
The Ontology-Violation Filter fills the relation blocklist where the verifier-fed
pathway did not, but the two are not redundant. The verifier rejects relations on
content grounds (hallucinated pairs, unsupported relations); the Filter rejects them
on structural grounds (type-triple violations against the ontology). On the held-out
cohort, the Filter flagged 49{,}734 ontology violations while the verifier continued
to reject content errors on a separate set of candidates. A deployment seeking
coverage of both error classes should run both sources.

\paragraph{(iii) Tagging rather than blocking preserves auditability.}
Automated removal of model output is contentious in clinical settings, because a
wrongly removed candidate is invisible by construction and therefore impossible to
audit. The DOWNGRADE-first discipline (Section~\ref{sec:design-downgrade}) tags
suspect candidates in the output stream instead of removing them, so downstream
consumers, clinicians, auditors, and training pipelines alike, see exactly what the
gate flagged and decide for themselves whether to act. A hard-blocking configuration
is available for deployments that prefer the gate's verdict to be final. We recommend
the tagging default in clinical contexts because a flagged candidate stays in the
audit trail whereas a blocked one disappears from it. This is an interpretive
recommendation following from the measurement discipline; the present work does not
directly compare the costs of a false block against those of a false allow in
clinical use.

\section{Limitations}
\label{sec:limits}

The empirical claims face four principal threats to validity, which we address in
turn: the narrow operating envelope of the NER-coverage gate
(Section~\ref{sec:limits-envelope}); the absence of a measured downstream precision
lift (Section~\ref{sec:limits-precision}); construct overlap between the gate's
design hypothesis test and its deployment evaluation
(Section~\ref{sec:limits-construct}); and single-corpus evaluation on PMC-Patients
(Appendix~\ref{app:single-corpus}, consolidated in
Section~\ref{sec:limits-appendix-pointer}). We disclose each so that readers and
reviewers can locate the bounds of the claims.

\subsection{Operating-Envelope Bounds of the NER-Coverage Gate}
\label{sec:limits-envelope}

The NER-coverage gate is selective in one operating regime, mid-spectrum coverage,
and degenerates in others. The per-category breakdown on the held-out cohort
(Section~\ref{sec:results-envelope}) identifies four regimes.

\begin{itemize}
    \item \textbf{Selective} (\texttt{diagnosis}, \texttt{risk\_factor}): the gate
    separates PASS from FAIL pairs. \texttt{diagnosis} (lift 17.13 at 95.7\%
    coverage) tags 73\% of FAIL pairs against 4\% of PASS pairs; \texttt{risk\_factor}
    (lift 2.16 at 53.7\% coverage) tags all FAIL pairs and 46\% of PASS pairs.
    \texttt{diagnosis} is selective only under full DBPM, where the
    Ontology-Violation Filter cleans the upstream coverage signal
    (Section~\ref{sec:discussion-principle}).

    \item \textbf{Categorical} (\texttt{outcome\_clinicalstatus},
    \texttt{outcome\_disposition}): at sparse coverage (4.4\% and 1.7\%) the gate
    tags nearly uniformly (lift 1.05 and 1.02), behaving as a category-level filter
    rather than a per-patient gate.

    \item \textbf{Saturated or no-tagging} (\texttt{history}, \texttt{observation},
    \texttt{symptoms}, \texttt{treatment}, \texttt{tests}): for \texttt{symptoms},
    \texttt{treatment}, and \texttt{tests}, coverage above 95\% means nearly every
    patient is covered and the gate allows almost everyone; for \texttt{history} and
    \texttt{observation}, the gate does not fire on this cohort. In both sub-regimes,
    the mechanism operates correctly but rarely activates.

    \item \textbf{No NER bucket} (\texttt{adverse\_event}, \texttt{outcome\_mortality}):
    the worker has no matching bucket, so the gate cannot apply and the difficulty-gate
    fallback tags PASS pairs only.
\end{itemize}

The headline path lift of 1.84 therefore averages behaviour across these regimes. A
deployment seeking uniform selectivity across all eleven categories would need to
address the saturated, sparse, and no-bucket regimes through schema closure
(Section~\ref{sec:future-schema}) and category-specific configuration.

\subsection{No Downstream Precision Lift Measured}
\label{sec:limits-precision}

The verifier is permissive at the scale tested: it passes essentially every relation
reaching the enriched output stage in both arms of both ablations. The precision
proxy delta on the held-out cohort is therefore $1.0000 / 1.0000$, and we cannot
claim that DBPM reduces downstream hallucinations as measured by the verifier's
decisions on enriched output.

What DBPM demonstrably does is upstream candidate suppression (the Ontology-Violation
Filter removes 913 relation candidates on the held-out cohort) and per-pair
selectivity tagging (the NER-coverage gate tags 6{,}101 PASS-side QA pairs against a
fully ungated baseline; Section~\ref{sec:results-downgrade}). These are measurable
effects on the pipeline's intermediate state, not measured improvements to the final
dataset's clinical correctness. Establishing that the algorithmic effects translate
into clinical-correctness gains requires human-rated true-positive evaluation on a
stratified sample of suppressed and tagged candidates, which is outside this paper's
scope (Section~\ref{sec:future-quality}) and reserved for separate dataset-quality
work.

\subsection{Construct Overlap in the Gate's Hypothesis Test}
\label{sec:limits-construct}

The development-cohort hypothesis test that motivated the NER-coverage gate used the
same construct, per-patient NER coverage against per-patient QA failure, that the
deployed gate now reads as its decision signal. The lift measured on that cohort is
therefore not a held-out estimate of the gate's generalisation.

We partially mitigate this by triangulating the path lift across four measurements
under different configurations and runs: the component-isolated ablation on the
held-out cohort gives 1.52, the cross-task full-configuration ablation gives 1.71,
the full-DBPM held-out ablation gives 1.84, and the development cohort gives a global
blended lift of 1.59.\footnote{The development-cohort entry is the \emph{global}
blended lift (1.59), not the path lift (1.89 on the nine mapped categories,
Section~\ref{sec:results-iterations}); we list it here because it is the figure the
hypothesis test reported, and we mark it as global to avoid conflating it with the
three path-lift measurements.} The three path-lift measurements span 1.52--1.84, and
every figure exceeds the 1.5 selectivity threshold. The full-DBPM configuration,
which we adopt as the primary result, sits at the top of this band because the
Ontology-Violation Filter removes ontology-violating relations that would otherwise
inflate the coverage signal, sharpening it when the full stack is active (the
\texttt{diagnosis} category, for instance, rises from anti-selective in isolation to
a lift of 17.1 under full DBPM; Section~\ref{sec:results-iterations}). Replication
across scales, configurations, and runs is not the same as held-out generalisation,
but the consistency of the path lift above threshold across every measurement
strengthens the mitigation. A genuine held-out validation would require a fresh
cohort drawn from a population the hypothesis test did not consume; we identify this
as future work (Section~\ref{sec:future-validation}).

\subsection{Further Limitations in Appendix~\ref{app:limits-supplementary}}
\label{sec:limits-appendix-pointer}

Two further limitations are retained in
Appendix~\ref{app:limits-supplementary} so the body stays focused on the three
principal bounds above:

\begin{itemize}
    \item Pattern-level cross-task propagation
    (Section~\ref{sec:design-crosstask}), previously flagged as not independently
    ablated, is now measured at 0.098\% of pipeline throughput, below the per-stage
    non-determinism floor (Appendix~\ref{app:crosstask}).

    \item Single-corpus evaluation: all results are measured on PMC-Patients;
    extension to electronic health record (EHR) corpora is reserved for future work
    (Section~\ref{sec:future-validation}), with the structural expectations under EHR
    distribution shift discussed in Appendix~\ref{app:single-corpus}.
\end{itemize}

\subsection{Scope of the Methodological Contribution}
\label{sec:limits-scope}

This paper makes claims about DBPM's algorithmic behaviour: which signal sources
populate which channels, which gate designs are selective, and which integrity checks
pass at scale. It does \emph{not} claim that DBPM's outputs improve the clinical
correctness of the resulting dataset, that the suppressed relations are genuinely
wrong in the clinical context of their source text, or that models trained on
DBPM-gated output outperform those trained on ungated output. These are
downstream-quality questions for separate evaluation work, deferred explicitly to
keep the present scope coherent.

\section{Future Work}
\label{sec:future}

We identify three future-work directions, ordered by how directly they extend the
paper's contributions. Each addresses a specific limitation from
Section~\ref{sec:limits}.

\subsection{Schema Closure for No-Bucket Categories}
\label{sec:future-schema}

Two categories (\texttt{adverse\_event}, \texttt{outcome\_mortality}) have no
matching NER bucket, so the NER-coverage gate cannot fire and the difficulty-gate
fallback tags PASS pairs only (Sections~\ref{sec:results-envelope}
and~\ref{sec:limits-envelope}). Extending the NER stage to emit
\texttt{Adverse\_Event} and \texttt{Outcome\_Mortality} entity types would let the
gate operate on these two categories on the same terms as the other nine. Closing the
schema this way would raise the \emph{global} blended lift toward the path lift, since
the two fallback categories would begin contributing selective tags rather than
zero-lift fallback tags; the realised gain depends on the NER yield in these
categories and the verifier's rejection distribution within them, and is an empirical
question for the closed-schema deployment. Schema closure is an engineering extension
that does not change the methodological contribution, but it is the most direct path
to broadening the gate's operating envelope.

\subsection{Dataset-Quality Evaluation}
\label{sec:future-quality}

This paper makes no claim that DBPM's suppressions and tags improve dataset quality
(Section~\ref{sec:limits-scope}). Establishing that would require a separate
programme on the pipeline's multi-task derivative of PMC-Patients: stratified human
review of suppressed relations and tagged QA pairs against clinical-correctness
criteria, together with downstream model-training studies on gated versus ungated
output. A stratified sample of suppressed relations across the six ontology-violation
patterns has been prepared as a release artefact; its evaluation is reserved for a
separate dataset-quality paper.

\subsection{Held-Out and EHR Validation}
\label{sec:future-validation}

The construct-overlap caveat (Section~\ref{sec:limits-construct}) is addressable by
validating the gate on a held-out PMC-Patients cohort drawn from a population the
hypothesis test did not consume. The single-corpus caveat
(Appendix~\ref{app:single-corpus}) is addressable by extending evaluation to
EHR-derived narratives, subject to the data-access constraints typical of EHR
research. The two validations are complementary: the first establishes that the gate
generalises on a similar distribution, the second that it generalises across a
documentation-style shift. Each warrants its own publication.

\section{Conclusion}
\label{sec:conclusion}

DBPM is an inference-time pattern-memory gating mechanism for a clinical NLP
pipeline, and this paper characterises its empirical behaviour at production scale
rather than its theoretical properties. The characterisation yielded four findings:
a verifier-fed accumulation pathway that fails at 167{,}034-patient production scale
(Section~\ref{sec:results-f1}); a verifier-independent ontology pathway that succeeds
where the verifier-fed pathway did not (Section~\ref{sec:results-ontology}); a
five-source dissection of the question-answering gate in which four sources fail for
structurally distinct reasons and one succeeds by aligning its signal with the
verifier's grounding criterion (Sections~\ref{sec:results-iterations}
and~\ref{sec:results-envelope}); and the structural principle that follows from the
dissection (Section~\ref{sec:discussion-principle}).

The gate's operating envelope is bounded. We characterise the bounds, saturated
coverage, sparse coverage, categorical regimes, and no-bucket categories
(Section~\ref{sec:limits-envelope}), and identify the extensions that would expand
them. We do not claim improvements to downstream clinical correctness: the effects we
measure are upstream candidate suppression and per-pair selectivity tagging, and
whether these translate into clinically correct dataset improvements is reserved for
separate dataset-quality work (Section~\ref{sec:future-quality}).

The five-source dissection is the contribution we expect to travel beyond this
pipeline. Practitioners designing inference-time gates for generator-verifier
architectures in any domain, clinical NLP, code generation, scientific extraction,
agentic tool use, can apply the structural principle to test a candidate signal source
before investing engineering effort in a mechanism that the four negative results here
suggest will not generalise. The principle is necessary, not sufficient:
signal-distributional properties further bound the operating regime. Both the
principle and the bounds emerge from work at production scale, where the failure modes
the paper documents become visible in a way they do not at benchmark scale.

We release the DBPM gating module, the ablation runner, the gate-decision graders,
and the held-out cohort identifier list. The release lets any reader reproduce the
source-by-source outcomes and the production-scale channel finding from the same
underlying data; the five signal sources are documented as a repository changelog.

\appendix
\section{Implementation, Operational, and Extension Details}
\label{app:main}

This appendix collects implementation engineering, operational configuration, and
extension-direction material referenced from the body of the paper.
Section~\ref{app:implementation} also serves as the in-code identifier reference
for the mechanisms that Section~\ref{sec:design} names by function: the body names
each flag, threshold, and method by what it does, and the mapping to source-level
identifiers is given here.

\subsection{Implementation and In-Code Identifier Reference}
\label{app:implementation}

DBPM is realised as a single Python class with $\mathcal{O}(1)$ amortised gate
cost. Read-side data structures are precomputed once per save into per-task hash
sets \texttt{\_block}, \texttt{\_downgrade}, \texttt{\_whitelist}, indexed by
signature; gate queries (Equation~\ref{eq:gate}) are constant-time set membership
with at most three lookups (whitelist, block, downgrade) plus a count comparison.
Write-side updates (Equations~\ref{eq:update-operator} and~\ref{eq:severity-update})
use a \texttt{\_write\_index} mapping signature keys to pattern dictionaries for
$\mathcal{O}(1)$ in-place modification.

State is persisted to a single JSON file (\texttt{bpm\_production.json}) under a
FileLock-protected atomic-write protocol: the in-memory state is serialised to a
temporary file and renamed over the canonical path under exclusive lock. To bound
I/O contention at the deployment's 64-way effective concurrency, saves are
batched: each worker process accumulates record events and triggers a save after
every 200 such events, then merges the on-disk state with its own using $\max$ for severity and
sum for counts. With 16 worker processes, this places save frequency at roughly one
merge per 12.5 events globally, below the FileLock acquisition rate threshold for
serialisation stalls.

\paragraph{Gate methods and signal sources.}
The three gate sites of the three-tier policy (Section~\ref{sec:design-gating}) are
exposed as \texttt{gate\_ner($\sigma$, NER)}, \texttt{gate\_relation($\sigma$, RE)},
and \texttt{gate\_qa(category, \ldots)}. The question-answering method retains the
legacy name \texttt{gate\_qa} for backward compatibility; the deployed
configuration runs on QAR-category inputs, and all selectivity claims are measured
against QAR outputs (Section~\ref{sec:background-pipeline}). The Ontology-Violation
Filter (Section~\ref{sec:design-ontology}) is \texttt{M3-RE} in the codebase: it
records through \texttt{bpm.record("re", payload, failure\_type="hard\_fail")} with
payload field \texttt{onto\_violation=True}, and populates the same
\texttt{\_re\_block} cache that \texttt{gate\_relation()} queries, introducing no
new gate.

\paragraph{Ablation flags.}
Six flags expose the design choices for ablation (Table~\ref{tab:app-flags}). The
four core flags toggle individual mechanisms; the gate-level flag
\texttt{M1\_QA\_GATE\_HARD\_BLOCK} controls whether the question-answering gate
emits BLOCK at all and defaults to $0$ (DOWNGRADE-first) in every measurement of
this paper (Section~\ref{sec:design-downgrade}); and the master switch
\texttt{BPM\_DISABLE} disables all DBPM gating and recording by flipping the four
core flags to OFF at worker initialisation. Because the master switch operates at
gate-invocation time rather than at configuration-serialisation time, the persisted
configuration block is byte-identical between a \texttt{BPM\_DISABLE=0} run and a
\texttt{BPM\_DISABLE=1} run; the architectural consequences of this are discussed
in Section~\ref{app:architecture}.

\begin{table}[h]
\centering
\caption{Module-level ablation flags and their effect. The default column gives the
value used in the deployed full-DBPM configuration; the master switch
\texttt{BPM\_DISABLE} overrides the four core flags when set.}
\label{tab:app-flags}
\small
\setlength{\tabcolsep}{6pt}
\renewcommand{\arraystretch}{1.3}
\begin{tabular}{lll}
\toprule
\textbf{Flag} & \textbf{Effect when OFF} & \textbf{Default} \\
\midrule
\texttt{BPM\_ENABLE\_CROSS\_TASK}     & No pattern-level cross-task propagation             & ON \\
\texttt{BPM\_ENABLE\_THREE\_TIER}     & Gate collapses to BLOCK / ALLOW                     & ON \\
\texttt{BPM\_ENABLE\_UNCERTAINTY}     & Ambiguous QAR verdicts treated as \texttt{success}  & ON \\
\texttt{BPM\_ENABLE\_SOURCE\_WEIGHTS} & All source weights set to $w_s = 1$                 & ON \\
\midrule
\texttt{M1\_QA\_GATE\_HARD\_BLOCK}    & QAR gate emits DOWNGRADE, never BLOCK               & $0$ \\
\texttt{BPM\_DISABLE}                 & All DBPM gating and recording active                & $0$ \\
\bottomrule
\end{tabular}
\end{table}

\paragraph{Thresholds and constants.}
Table~\ref{tab:app-constants} maps the concept names used in
Section~\ref{sec:design-params} to their in-code identifiers and deployed values.
The learning rates and decay factors are given in Table~\ref{tab:learning-rates}
and the body of Section~\ref{sec:design-params}; the source weights are in
Table~\ref{tab:source-weights}. The question-answering fail-rate threshold of the
third and fourth signal sources (Equations~\ref{eq:m1v5-failrate}
and~\ref{eq:m1v6-difficulty}) is \texttt{M1V5\_FAIL\_THRESHOLD}, set to $0.30$ on
the development cohort and $0.60$ on the held-out 5{,}000-patient cohort; the
difficulty threshold of the fourth source is \texttt{M1V6\_DIFFICULTY\_THRESHOLD}
$=0.30$.

\begin{table}[h]
\centering
\caption{Concept-to-identifier mapping for the thresholds and constants of
Section~\ref{sec:design-params}. Per-task values are given as (NER, RE, QAR).}
\label{tab:app-constants}
\small
\setlength{\tabcolsep}{6pt}
\renewcommand{\arraystretch}{1.3}
\begin{tabular}{lll}
\toprule
\textbf{Concept (body)} & \textbf{In-code identifier} & \textbf{Deployed value} \\
\midrule
Block threshold $\theta_t^{\mathrm{block}}$    & \texttt{BLOCK\_THRESHOLDS[t].min\_severity}  & $(0.65,\,0.70,\,0.60)$ \\
Minimum support $n_t^{\min}$ (block)           & \texttt{BLOCK\_THRESHOLDS[t].min\_count}     & $(3,\,2,\,3)$ \\
Downgrade threshold $\theta_t^{\mathrm{down}}$ & downgrade band lower bound                   & $0.40$ (all tasks) \\
Minimum support (downgrade)                    & --                                           & $1$ (all tasks) \\
Storage cap $C_t$                              & per-task pattern cap                         & $(200,\,300,\,150)$ \\
Severity-zone guard                            & cross-task severity ceiling                  & $0.55$ \\
Propagation cap                                & \texttt{cross\_task\_hits} ceiling           & $3$ \\
QAR fail-rate threshold                        & \texttt{M1V5\_FAIL\_THRESHOLD}               & $0.30$ dev / $0.60$ held-out \\
QAR difficulty threshold                       & \texttt{M1V6\_DIFFICULTY\_THRESHOLD}         & $0.30$ \\
DOWNGRADE confidence multiplier                & calibrated-confidence scale                  & $0.70$ \\
\bottomrule
\end{tabular}
\end{table}

\begin{sloppypar}
\paragraph{Memory state and persistence fields}
Each blocklist entry stores the tuple $(\sigma_{\mathrm{sev}}, c,
\tau_{\mathrm{last}}, \kappa)$ of Section~\ref{sec:design-severity} under
\texttt{tasks[t].patterns} in \texttt{bpm\_production.json}, with the cross-task hit
counter $\kappa$ held in the field \texttt{cross\_task\_hits}; whitelist entries are
stored under \texttt{whitelist[t]} as $(\sigma_{\mathrm{conf}}, c,
\tau_{\mathrm{last}})$. Rejection events are logged to
\texttt{universal\_rejections.jsonl} as the tuple
$(\mathrm{uid}, \mathrm{stage}, \mathrm{entity}, \mathrm{reason}, \mathrm{conf})$ of
Section~\ref{sec:background-verdict-stream}. The deterministic question-to-category
map consulted by the NER-coverage gate is \texttt{QA\_TEMPLATES\_GEN}; the
ontology and alias tables of Equation~\eqref{eq:ontology} are
\texttt{ENTITY\_RELATION\_PRIORS} (29 admissible type-triples) and
\texttt{RELATION\_ALIAS} (32 normalisation entries), both fixed at worker load
time. The per-pair gate verdict is written to the output field \texttt{dbpm\_gate},
the cross-task recursion guard of Algorithm~\ref{alg:dbpm-update} is the parameter
\texttt{is\_cross\_task}, and the inactive self-critique source of
Table~\ref{tab:source-weights} is gated by \texttt{SELFVERIF\_ENABLE}, which
defaults to OFF.
\end{sloppypar}

\subsection{Verifier Verdict Semantics}
\label{app:verifier-semantics}

A design choice of the surrounding pipeline is worth noting because it shapes the
verdict distribution DBPM consumes. The verifier is conservative-pass on binary
tasks: for NER, only an explicit ``NO'' yields FAIL, and ambiguous outputs default
to PASS. For RE, ambiguous outputs default to \texttt{PASS\_WEAK}, the middle of the
graded scale. For QAR, a single ambiguous output triggers a three-vote ensemble with
tie-break to PASS.

These conservative defaults reduce false-rejection on borderline candidates and
produce a verdict distribution skewed toward acceptance, which is the appropriate
prior in clinical extraction where over-rejection systematically loses recall on
rare entities. DBPM's gating policy (Equation~\ref{eq:gate}) compensates by
requiring $n_t^{\min}$ corroborating failures before a BLOCK fires, so the
conservative-pass verifier and the support-count threshold work together rather than
at cross-purposes.

\subsection{Hardware and Software Configuration}
\label{app:hardware}
Table~\ref{tab:config} consolidates every fixed setting of the measurement, the
models and their decoding parameters, the gating constants, the evaluation cohort,
the ablation protocol, and the hardware, in a single reference; the remainder of this
subsection expands the serving and concurrency configuration in detail.

\begin{table}[H]
\centering
\caption{Complete experimental configuration. All settings were fixed before
analysis and are reported here so the measurement and downstream experiments can be
reproduced exactly. The gating constants summarised below are stated in full, with
their in-code identifiers, in Tables~\ref{tab:app-flags} and~\ref{tab:app-constants};
learning rates and source weights are in Tables~\ref{tab:learning-rates}
and~\ref{tab:source-weights}.}
\label{tab:config}
\footnotesize
\renewcommand{\arraystretch}{1.0}
\begin{tabular}{@{} l p{8.5cm} @{}}
\toprule
\textbf{Component} & \textbf{Setting} \\
\midrule
\multicolumn{2}{@{}l}{\textbf{\textit{Pipeline and corpus}}} \\
Source corpus              & PMC-Patients (167{,}034 patient narratives) \\
Generator model            & Llama-3.3-70B (instruction-tuned) \\
Verifier model             & MMed-Llama-3.1-70B \\
Task buckets               & 11 total; 3 DBPM-gated (NER, RE, QAR) \\
Serving framework          & vLLM, continuous batching, tensor parallelism TP $= 2$ per model \\
\addlinespace[0.3em]
\multicolumn{2}{@{}l}{\textbf{\textit{Decoding (generator and verifier)}}} \\
Context window             & 8192 tokens (\texttt{--max-model-len}) \\
Max concurrent sequences   & 256 (\texttt{--max-num-seqs}) \\
GPU memory utilisation     & 0.90 (\texttt{--gpu-memory-utilization}) \\
Precision                  & bfloat16 (\texttt{--dtype bfloat16}); chunked prefill enabled \\
Sampling temperature       & 0.0 (greedy) for all gated-task generation and verification \\
Top-$p$ / top-$k$          & Not applicable at temperature 0 (top-$p$ 0.9 engaged only if temperature $> 0$); no top-$k$ \\
Max generation tokens      & Per stage: 1500 (RE/JSON extraction), 160 (QAR), 60 (verifier verdict) \\

\addlinespace[0.3em]
\multicolumn{2}{@{}l}{\textbf{\textit{DBPM gating}}} \\
Block thresholds $\theta_t^{\mathrm{block}}$    & 0.65 / 0.70 / 0.60 (NER / RE / QAR) \\
Downgrade threshold $\theta_t^{\mathrm{down}}$  & 0.40 (all tasks) \\
Minimum support $n_t^{\min}$ (block)            & 3 / 2 / 3 (NER / RE / QAR); 1 for downgrade \\
Storage caps $C_t$                              & 200 / 300 / 150 (NER / RE / QAR) \\
Wall-clock half-lives $h_t$                     & 5 / 10 / 10 days (NER / RE / QAR) \\
In-run decay $\rho_v$                           & 0.99 (hard\_fail), 0.985 (soft\_downgrade) \\
Severity-zone guard                             & 0.55 (cross-task severity ceiling) \\
Propagation cap                                 & 3 cross-task hits per signature \\
DOWNGRADE confidence multiplier                 & 0.70 \\
QAR fail-rate threshold                         & 0.30 (development) / 0.60 (held-out) \\
QAR difficulty threshold                        & 0.30 \\
Measurement mode                                & DOWNGRADE-first (\texttt{M1\_QA\_GATE\_HARD\_BLOCK} $= 0$) \\
\addlinespace[0.3em]
\multicolumn{2}{@{}l}{\textbf{\textit{Evaluation cohort}}} \\
Held-out cohort            & 5{,}000 patients, stratified random subsample \\
Stratification             & Age bucket $\times$ gender $\times$ document-length tertile (24 non-empty strata) \\
Tertile cut points         & T1 $\leq 1{,}931 <$ T2 $\leq 3{,}121 <$ T3 characters \\
Sampling                   & Floor allocation then proportional residual; uniform random without replacement \\
Seed                       & Fixed (\texttt{random.Random(42)}) \\
Production cross-check      & Full corpus, 167{,}034 patients, deployed DBPM-full configuration \\
\addlinespace[0.3em]
\multicolumn{2}{@{}l}{\textbf{\textit{Ablation protocol}}} \\
Paired ablations           & 3 (Ontology Filter; NER-coverage gate; full DBPM master switch) \\
Cross-task ablation        & 1 separate paired ablation (reported in Appendix~\ref{app:crosstask}) \\
Selectivity metric         & Lift $= P(\mathrm{tag}\mid\mathrm{FAIL}) / P(\mathrm{tag}\mid\mathrm{PASS})$ \\
Significance               & Delta-method 95\% CI on $\log$ lift; $H_0\!:\mathrm{lift}=1$ \\
\addlinespace[0.3em]
\multicolumn{2}{@{}l}{\textbf{\textit{Hardware}}} \\
GPU                        & 4 $\times$ NVIDIA H200 (141\,GB HBM3e each) \\
System                     & 64 CPU cores, 900\,GB system memory \\
Scheduler                  & SLURM (Supercomputing Wales Falcon) \\
Persistence                & Single JSON file, FileLock atomic writes, 200-event save batches \\
\bottomrule
\end{tabular}
\end{table}

Experiments run on a single SLURM-managed compute node: $4 \times$ NVIDIA H200 GPUs
(141\,GB HBM3e each), 64 CPU cores, 900\,GB system memory. SLURM allocations are
72-hour wall-clock with a 70-hour Python-side budget and a 2-hour grace window for
graceful shutdown.

The pipeline is deployed as two independent vLLM OpenAI-compatible API servers, the
generator on port 8000 and the verifier on port 8001, each occupying two GPUs under
tensor parallelism $\mathrm{TP} = 2$:

\begin{itemize}
    \item \textbf{Generator $G$.} Llama-3.3 70B (instruction-tuned) on GPUs 0--1,
    configured with \texttt{--max-model-len 8192}, \texttt{--max-num-seqs 256},
    \texttt{--gpu-memory-utilization 0.90}, \texttt{--dtype bfloat16}, and
    \texttt{--enable-chunked-prefill}.
    \item \textbf{Verifier $V$.} MMed-Llama-3.1 70B \citep{qiu2024mmed} on GPUs
    2--3, with identical vLLM parameters.
\end{itemize}

The orchestrator spawns 16 worker processes via Python \texttt{multiprocessing}.
Each worker maintains four concurrent patient coroutines, giving 64-way effective
concurrency that vLLM's continuous batching aggregates into efficient GPU batches.
HTTP traffic between worker coroutines and the two vLLM servers is asynchronous
(\texttt{aiohttp}), with per-call timeouts of 180\,s and three retry attempts.
Running independent generator and verifier servers, rather than tensor-parallelising
a single model across all four GPUs, avoids the larger cross-GPU synchronisation that
a four-way split would incur. When one model is split across GPUs, the partial results
each GPU computes must be summed and redistributed to every GPU before computation
continues, an \emph{AllReduce} collective operation that fires on every token; keeping
each model within its own pair of GPUs confines this synchronisation to two GPUs rather
than four. This is the configuration used for the production runs. The save-batching and FileLock
protocol that bounds persistence contention at this concurrency is described in
Section~\ref{app:implementation}.

\subsection{Checkpoint-and-Resume Protocol}
\label{app:resume}

Because the full 167{,}034-patient corpus exceeds a single SLURM allocation, the
pipeline implements a checkpoint-and-resume protocol. The orchestrator detects
pre-existing \texttt{multitask\_data\_enriched.jsonl} output and skips
already-processed UIDs, opening output files in append mode so no record is
overwritten. The SLURM script monitors Python's exit code and treats codes 124
(timeout fired with SIGTERM), 130 (SIGINT-handled), 137 (SIGKILL escalation), and
143 (clean SIGTERM exit) as resubmit-eligible, computing the remaining patient count
and re-\texttt{sbatch}-ing the job automatically.

DBPM state survives across job submissions because its JSON file is written to the
shared output directory (\texttt{OUTPUT\_DIR/bpm\_production.json}), not to
node-local \texttt{/tmp}.

\subsection{Reproducibility}
\label{app:reproducibility}

The DBPM gating module and the analysis artefacts required to reproduce the paper's
quantitative claims are released under an open-source license (see Code
availability). The release includes:

\begin{itemize}
    \item the standalone DBPM gating module (\texttt{dbpm.py}), with all numeric
    constants stated in Section~\ref{sec:design-params} as class attributes;
    \item the selectivity grader (\texttt{grade\_m1v7\_selectivity.py}), the
    aggregate-provenance script (\texttt{aggregate\_dbpm\_provenance.py}), and the
    167K cross-check (\texttt{crosscheck\_F1\_167k\_blocklist.py});
    \item the genericised ablation runner (\texttt{run\_ablation.sh}), which
    documents the full-ON vs full-OFF protocol; and
    \item the held-out 5{,}000-patient and development-cohort identifier lists.
\end{itemize}

The surrounding extraction pipeline (the worker, orchestration, and model-serving
layers) is not included; the released module is the gating logic the paper
characterises, and the cohort lists plus aggregate counts permit reproduction of the
reported selectivity, lift, and production-scale channel findings. The five
question-answering signal sources are documented as a changelog in the repository
rather than shipped as patch scripts against the withheld worker.

\subsection{Architectural Disclosures}
\label{app:architecture}

DBPM gates three of the eleven pipeline task buckets (NER, RE, QAR;
Section~\ref{sec:background-pipeline}). The summary task has a reserved threshold in
\texttt{BLOCK\_THRESHOLDS["summary"]} but no consumer gate. Medication extraction,
temporal-event extraction, active-risk identification, risk-state-machine
derivation, risk-based recommendations, risk-grounded QAR, and visualisation-payload
assembly are entirely outside the DBPM gating system. Extension to these tasks is
reserved for future work.

DOWNGRADE-first measurement discipline (Section~\ref{sec:design-downgrade}) is
applied to new, untested signal channels: the third signal source onward of the
question-answering gate, and the Ontology-Violation Filter channel. Legacy
production-validated channels (the named-entity gate over \texttt{\_ner\_block}, and
the legacy \texttt{\_re\_block} check in \texttt{gate\_relation}) retain hard-BLOCK
semantics as a deliberate design choice, since they have empirical support from the
project's deployment history.

The byte-identical configuration footprint of the \texttt{BPM\_DISABLE} master
switch, noted in Section~\ref{app:implementation}, has one useful empirical
consequence: because the switch acts at gate-invocation time, the OFF arm of the
combined ablation (Section~\ref{sec:results-downgrade}) records zero DOWNGRADE tags
across the full 5{,}000-patient cohort, confirming that the switch produces the
expected gate inactivity rather than a silently altered configuration.

\subsection{Pattern-Level Cross-Task Propagation: Measured Contribution}
\label{app:crosstask}

The pattern-level cross-task propagation mechanism described in
Section~\ref{sec:design-crosstask} is a persistence-layer component active in all
DBPM runs reported in Section~\ref{sec:results}. We isolate its independent
contribution via a single paired ablation: ON arm
\texttt{BPM\_ENABLE\_CROSS\_TASK = 1} (the deployed default) versus OFF arm
\texttt{BPM\_ENABLE\_CROSS\_TASK = 0}, with all other DBPM mechanisms
(Ontology-Violation Filter, NER-coverage gate, three-tier gating, source weights)
held at their default ON settings. Both arms use the held-out 5{,}000-patient
cohort.

\paragraph{Result.}
The cross-task ablation produces a combined pair-count delta of 280 across the three
gated tasks ($\mathrm{NER}~\Delta = -232$, $\mathrm{RE}~\Delta = +41$,
$\mathrm{QAR}~\Delta = -7$) against 287{,}350 total pipeline pairs in the ON arm,
yielding a contribution of 0.098\% of pipeline throughput. This is below the
0.5\% per-stage non-determinism floor of the pipeline (vLLM continuous batching plus
the dual-model verifier introduce $\sim 0.1$--$0.3\%$ per-stage run-to-run variance
on identical-cohort runs; see Appendix~\ref{app:resume}). The path lift in the ON arm
is 1.71, within the 1.52--1.84 empirical band documented in
Section~\ref{sec:limits-construct}.

\paragraph{Instrumentation note.}
The worker's \texttt{\_event\_counters["cross\_task\_propagations"]} global counter
reported zero events in this ablation, while 24 NER patterns in the persisted
\texttt{bpm\_production.json} recorded \texttt{cross\_task\_hits} $> 0$. The
discrepancy reflects an instrumentation gap in the worker's event-counter update
paths: cross-task propagation through new-signature creation increments the
per-pattern counter but not the global event counter. The per-pattern evidence
confirms the mechanism fired, and the pair-count delta is the authoritative
measurement of mechanism contribution at the pipeline-output level. We disclose this
for transparency; the substantive claim is the 0.098\% pair-delta, which is
independent of the global event-counter discrepancy.

\paragraph{Interpretation.}
The mechanism's safeguards (severity-zone guard at 0.55, propagation cap
$\kappa \geq 3$, source weights $w \in [0.2, 0.4]$) bound its impact below the
per-stage non-determinism floor at the held-out 5K scale. This is consistent with the
architectural intent of the mechanism: pattern-level cross-task propagation is
designed to push severity into the DOWNGRADE band without crossing into BLOCK on
cross-task evidence alone (Section~\ref{sec:design-gating}, invariant 3), and at this
scale the safeguards bind before any output-level effect accumulates. The result
closes the independent-ablation limitation flagged in earlier drafts of this paper.

\subsection{Single-Corpus Evaluation}
\label{app:single-corpus}

All evaluation in this paper uses PMC-Patients \citep{zhao2023large} as the source
corpus. PMC-Patients is the established public benchmark for clinical narrative
extraction at the sub-100K-patient-record scale, and using it permits direct
comparison to prior LLM-driven clinical NLP work. However, the corpus consists of
case reports from PubMed Central full-text articles, which differ in documentation
style from electronic health records (EHRs) used in real clinical settings: case
reports are authored for academic publication, are typically reviewed by peers
before release, and are written with denotative precision intended to support
reproducibility. EHR narratives are authored under clinical time pressure, contain
abbreviations and shorthand specific to local institutional cultures, and reflect
dictation artefacts not present in published case reports.

The empirical findings of Section~\ref{sec:results}, specifically the
diffuse-rejection distributional shape that produced the production-scale
channel finding (Section~\ref{sec:results-f1}) and the per-category operating
envelope of the NER-coverage gate (Section~\ref{sec:results-envelope}), may shift on
EHR-derived corpora. Extension to EHR data is identified as future work
(Section~\ref{sec:future-validation}).

\subsection{Answerability Cascade for Saturated and Sparse-Coverage Categories}
\label{app:answerability}

Five QAR categories on the held-out cohort exhibit saturated NER coverage
(\texttt{symptoms}, \texttt{treatment}, \texttt{tests}, \texttt{observation},
\texttt{history}); the NER-coverage gate rarely fires for these because nearly every
patient is covered. One category (\texttt{outcome\_clinicalstatus}) exhibits sparse
NER coverage, so the gate fires categorically rather than per-patient. A
complementary signal source, for example a distilled small-language-model
answerability check that fires when NER coverage is uninformative, could extend the
working operating envelope to these categories while preserving the deterministic
predicate structure of the present design.

This extension is conditional on its own evaluation establishing that the additional
signal is selective; we do not commit to it working at this stage. We identify it as
the most natural next-step extension for deployments seeking uniform per-pair
selectivity across all categories.

\subsection{Limitations Addressed in This Appendix}
\label{app:limits-supplementary}

This section consolidates the appendix-resident limitation disclosures referenced
from Section~\ref{sec:limits-appendix-pointer}:
\begin{itemize}
    \item Pattern-level cross-task propagation (Appendix~\ref{app:crosstask}):
    measured contribution 0.098\% of pipeline throughput, below the per-stage
    non-determinism floor.
    \item Single-corpus evaluation (Appendix~\ref{app:single-corpus}): all empirical
    findings are measured on PMC-Patients; EHR distribution-shift is addressed in
    Section~\ref{sec:future-validation}.
\end{itemize}

\printcredits

\section*{Code availability}
The DBPM gating module, the grader scripts that reproduce the reported selectivity
and lift figures, and the cohort identifier lists are openly available on GitHub at
\url{https://github.com/Ali-Lazem/DBPM} and permanently archived on Zenodo at
\url{https://doi.org/10.5281/zenodo.21038023}. The surrounding extraction pipeline is
not included; the repository provides the gating logic and the artefacts required to
reproduce the paper's quantitative claims.

\section*{Data availability}
The cohort identifier lists used for the reported ablations are included in the code
repository and index the publicly available PMC-Patients corpus
\citep{zhao2023large}. The aggregate gating-outcome counts that reproduce the
reported figures are provided in the repository (counts and rates only; no clinical
text). The derived extraction corpus is not released.

\section*{Acknowledgements}
We acknowledge the services of Supercomputing Wales and the Bangor eResearch Team,
including the support provided by Dr. Ade Fewings.

\bibliographystyle{cas-model2-names}

\bibliography{cas-refs}

\vskip3pt

\bio{}
\textbf{Ali Hussein Lazem} received the B.Sc. in Computer Science from the University of Thi-Qar, Iraq, and the M.Sc. in Artificial Intelligence from Troy University, Alabama, USA. He is currently a Ph.D. candidate in the School of Computer Science and Engineering at Bangor University, United Kingdom. His research focuses on the production-scale deployment of large language models for clinical natural language processing, with broader interests in generative AI, computer vision, and machine learning.
\endbio

\vskip3pc

\bio{}
\textbf{William J. Teahan} is a Senior Lecturer and Postgraduate Research Director in the School of Computer Science and Engineering at Bangor University, United Kingdom. His research spans natural language processing, text compression, machine learning, and agent-based systems, with a long-standing interest in statistical models of language and structured knowledge representation. He has contributed to multidisciplinary projects across computer science, biological sciences, and environmental sciences, with collaborations spanning multiple countries. His roles at Bangor include postgraduate supervision, curriculum development, and the coordination of research programmes within the School. He has supervised numerous doctoral students whose work has informed computational approaches to language modelling and applied artificial intelligence.
\endbio
\clearpage

\end{document}